\documentclass{egpubl}
 
\JournalSubmission    
\usepackage[T1]{fontenc}
\usepackage{dfadobe}  

\biberVersion
\BibtexOrBiblatex
\usepackage[backend=biber,bibstyle=EG,citestyle=alphabetic,backref=true]{biblatex} 
\electronicVersion
\PrintedOrElectronic

\ifpdf \usepackage[pdftex]{graphicx} \pdfcompresslevel=9
\else \usepackage[dvips]{graphicx} \fi

\usepackage{egweblnk} 

\usepackage{amsmath}
\usepackage{amssymb}
\usepackage{bm}
\usepackage{booktabs, multirow} 

\DeclareSymbolFont{letters}     {OML}{cmm} {m}{it}
\DeclareMathAlphabet      {\mathbf}{OT1}{cmr}{bx}{n}

\title[3D Generative Model Latent Disentanglement via Local Eigenprojection]%
      {3D Generative Model Latent Disentanglement \\ via Local Eigenprojection}

\author[S. Foti et al.]{
    \parbox{\textwidth}{\centering 
        Simone Foti$^{1}$\orcid{0000-0003-3207-1965},
        Bongjin Koo$^{1,2}$\orcid{0000-0002-3611-4988},
        Danail Stoyanov$^{1}$\orcid{0000-0002-0980-3227}, and
        Matthew J. Clarkson$^{1}$\orcid{0000-0002-5565-1252}
    } \\ 
    {\parbox{\textwidth}{\centering 
        $^1$University College London, London, UK\\
        $^2$University of California, Santa Barbara, Santa Barbara, USA
    }}
}

\begin{document}

\teaser{
    \centering
    \includegraphics[width=\textwidth]{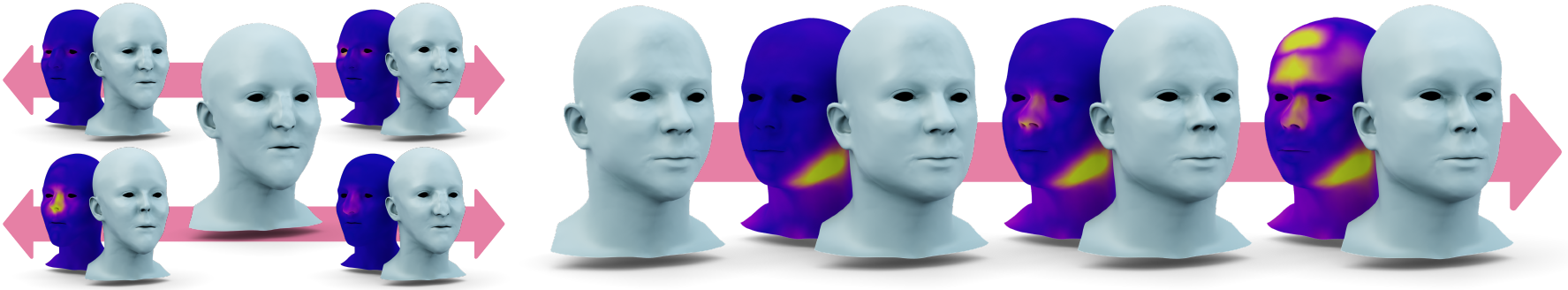}
    \caption{Shape generation and editing of two subjects randomly generated with LED-VAE, which is one of the proposed local eigenprojection disentangled models. \textit{Left:} effects caused on the generated shapes by traversing two arbitrary latent variables controlling the eyes and nose of the first random subject. \textit{Right:} example of shape editing performed manipulating the latent variables controlling jaw, nose, and forehead of the second subject. The latent manipulations are performed with a GUI that allows the manual modification of the latent variables, but random per-attribute modifications can also be performed. The edited shapes are always paired with their corresponding displacement map highlighting the shape differences from the initial model. }
    \label{fig:teaser}
}

\maketitle
\begin{abstract}
    Designing realistic digital humans is extremely complex. Most data-driven generative models used to simplify the creation of their underlying geometric shape do not offer control over the generation of local shape attributes. In this paper, we overcome this limitation by introducing a novel loss function grounded in spectral geometry and applicable to different neural-network-based generative models of 3D head and body meshes. Encouraging the latent variables of mesh variational autoencoders (VAEs) or generative adversarial networks (GANs) to follow the local eigenprojections of identity attributes, we improve latent disentanglement and properly decouple the attribute creation. Experimental results show that our local eigenprojection disentangled (LED) models not only offer improved disentanglement with respect to the state-of-the-art, but also maintain good generation capabilities with training times comparable to the vanilla implementations of the models. Our code and pre-trained models are available at \href{https://github.com/simofoti/LocalEigenprojDisentangled}{github.com/simofoti/LocalEigenprojDisentangled}.



\end{abstract}  

\section{Introduction}
    \label{sec:intro}
    In recent years digital humans have become central elements not only in the movie and video game production, but also in augmented and virtual reality applications. With a growing interest in the metaverse, simplified creation processes of diverse digital humans will become increasingly important. 
    These processes will benefit experienced artists and, more importantly, will democratise the character generation process by allowing users with no artistic skills to easily create their unique avatars.
    Since digitally sculpting just the geometric shape of the head of a character can easily require a highly skilled digital artist weeks to months of work~\cite{gruber2020interactive}, many semi-automated avatar design tools have been developed. Albeit simpler and faster to use, they inherit the intrinsic constraints of their underlying generative models~\cite{foti2021swapdisentangled}. Usually based upon blendshapes~\cite{loper2015smpl, osman2020star, tena2011interactive}, principal component analysis (PCA)~\cite{blanz1999morphable, ploumpis2019combining, li2017learning}, variational autoencoders (VAEs)~\cite{ranjan2018generating, gong2019spiralnet++, aumentado2019geometric, cosmo2020limp}, or generative adversarial networks (GANs)~\cite{cheng2019meshgan, gecer2020synthesizing, li2020learning, abrevaya2019decoupled}, these models are either limited in expressivity or they cannot control the creation of local attributes. 
    Considering that deep-learning-based approaches, such as VAEs and GANs, offer superior representation capabilities with a reduced number of parameters and that they can be trained to encourage disentanglement, we focus our study on these models. 
    
    By definition~\cite{bengio2013representation, higgins2016beta, kim2018disentangling}, with a disentangled latent representation, changes in one latent variable affect only one factor of variation while being invariant to changes in other factors. This is a desirable property to offer control over the generation of local shape attributes. However, latent disentanglement remains an open problem for generative models of 3D shapes~\cite{aumentado2019geometric} despite being a widely researched topic in the deep learning community~\cite{higgins2016beta, kim2018disentangling, kulkarni2015deep, esmaeili2019structured, ding2020guided, wang2021self, rhodes2021local}. Most research on latent disentanglement of generative models for the 3D shape of digital humans addresses the problem of disentangling the pose and expression of a subject from its identity~\cite{aumentado2019geometric, aumentado2021disentangling, cosmo2020limp, abrevaya2019decoupled, zhang2020learning, lombardi2021latenthuman, zheng2022imface}, but none of these works is able to provide disentanglement over the latent variables controlling the local attributes characterising the identity. Some control over the generation of local attributes was achieved for generative models of 3D furniture by leveraging complex architectures with multiple encoders and decoders independently operating on different furniture parts~\cite{nash2017shape, yang2020dsm, roberts2021lsd}. In contrast, \cite{foti2021swapdisentangled} recently proposed a method to train a single VAE while enforcing disentanglement among sets of latent variables controlling the identity of a character. This approach allows their Swap Disentangled VAE (SD-VAE) to learn a more disentangled, interpretable, and structured latent representation for 3D VAEs of bodies and heads. However, although \cite{foti2021swapdisentangled} disentangles subsets of latent variables controlling local identity attributes, variables within each set can be entangled and not orthogonal. In addition, their curated mini-batching procedure based on attribute swapping is applicable only to autoencoder-based architectures and it significantly increases the training duration. In this work, we aim at overcoming these limitations by leveraging spectral geometry to achieve disentanglement without curating the mini-batching. In particular, we encourage the latent representation of a mesh to equal the most significant local eigenprojections of signed distances from the mean shape of the training data. 
    Since the eigenprojections are computed using the eigenvectors of combinatorial Laplacian operators, we require meshes to be in dense point correspondence and to share the same topology. This is a standard requirement for most of the traditional~\cite{blanz1999morphable, booth20163d, dai2020statistical, gruber2020interactive, loper2015smpl, osman2020star, ploumpis2019combining, ploumpis2020towards} and neural-network-based~\cite{foti2020intraoperative, foti2021swapdisentangled, gong2019spiralnet++, ranjan2018generating, zhou2020fully, yuan2020mesh} generative models, which not only simplifies the shape generation process, but also the definition of other digital humans' properties that will be automatically shared by all the generated meshes (e.g., UV maps, landmarks, and animation rigs). 
    
    To summarise, the key contribution of this work is the introduction of a novel local eigenprojection loss, which is able to improve latent disentanglement among variables controlling the generation of local shape attributes contributing to the characterisation of the identity of digital humans. Our method improves over SD-VAE 
    by enforcing orthogonality between latent variables and avoiding the curated mini-batching procedure, thus significantly reducing the training times. 
    In addition, we demonstrate the flexibility and disentanglement capabilities of our method on both VAEs and GANs.

\section{Related Work}
    \label{sec:related_work}
    
    \subsection{Generative Models.}
        \label{sec:related_work_gen_models}
        Blendshapes are still widely adopted for character animation or as consumer-level avatar design tools because, by linearly interpolating between a predefined set of artistically created shapes, the blend-weights can be easily interpreted~\cite{lewis2014practice}. However, to compensate for the limited flexibility and diversity of these models, large amounts of shapes are required. This makes the models very large and only a limited number of shapes can be used in most practical applications. 
        An alternative approach capable of offering more flexibility is to build models relying on principal component analysis (PCA)~\cite{blanz1999morphable, egger20203d}. 
        These data-driven models are able to generate shapes as linear combinations of the training data, but the variables controlling the output shapes are related to statistical properties of the training data and are difficult to interpret. In recent years, PCA-based models have been created from large number of subjects. For example, \textsc{Lsfm}~\cite{booth20163d} and \textsc{Lyhm}~\cite{dai2020statistical} were built collecting scans from $10,000$ faces and $1,212$ heads respectively. The two models were later combined in \textsc{Uhm}~\cite{ploumpis2019combining}, which was subsequently enriched with additional models for ears, eyes, teeth, tongue, and the inner-mouth~\cite{ploumpis2020towards}.
        Also, \cite{gruber2020interactive} combined multiple PCA models, but they were controlling different head regions and an anatomically constrained optimisation was used to combine their outputs and thus create an interactive head sculpting tool. PCA-based models of the body were also combined with blendshapes in \textsc{Smpl}~\cite{loper2015smpl} and \textsc{Star}~\cite{osman2020star}, which were trained with $3,800$ and $14,000$ body scans respectively. PCA-based models generally trade the amount of fine details they can represent with their size. The advent of geometric deep learning techniques brought a new set of operators making possible the creation of neural network architectures capable of processing 3D data such as point-clouds and meshes. \cite{ranjan2018generating} introduced the first VAE for the generation of head meshes. In its comparison against PCA, the VAE model used significantly fewer parameters and exhibited superior performances in generalisation, interpolation, and reconstruction. This pioneering work was followed by many other autoencoders which differed from one another mostly by their application domain and the mesh operators used in their architecture~\cite{litany2018deformable, foti2020intraoperative, yang2018foldingnet, zhou2020fully, gong2019spiralnet++, dai2019pointae, tan2022mesh, bouritsas2019neural}.   
        These mesh operators were used also for generative models based on GAN architectures \cite{olivier2021facetunegan, cheng2019meshgan}, but they appear to be less frequent than their VAE counterparts.
        Most GAN architectures operate in the image domain by representing 3D shapes in a UV space~\cite{moschoglou20203dfacegan, li2020learning}.
    
    
    \subsection{Latent Disentanglement.}
        Most research on latent disentanglement is performed on generative models of images~\cite{kumar2017variational, kim2018disentangling, kulkarni2015deep, esmaeili2019structured, ding2020guided, rhodes2021local, wang2021self}. The $\beta$-VAE~\cite{higgins2016beta} is probably the simplest model used to improve disentanglement in a VAE. Other simple methods that leverage statistical properties and do not require supervision over the generative factors are for instance the DIP-VAEs~\cite{kumar2017variational} and the FactorVAE~\cite{kim2018disentangling}. All methods above were re-implemented to operate on meshes by~\cite{foti2021swapdisentangled}, but they did not report good levels of disentanglement with respect to the identity attributes. In the 3D realm, there are currently two prominent streams of research: the one disentangling the identity from the pose or expression of digital humans~\cite{aumentado2019geometric, aumentado2021disentangling, cosmo2020limp, zhang2020learning, zhou2020unsupervised, tatro2021unsupervised, jiang2019disentangled, huang2021arapreg,otberdout2022sparse}, and the stream attempting to disentangle parts of man-made objects~\cite{yang2020dsm, nash2017shape, li2021editvae, roberts2021lsd}. In both cases, the proposed solutions require complex architectures. In addition, in the former category, current state-of-the-art methods do not attempt to disentangle identity attributes. The latter category appears better suited for this purpose, but the type of generated shapes is substantially different because the generation of object parts needs to consider intrinsic hierarchical relationship, and surface discontinuities are not a problem.
        More similar to ours, is the method recently proposed by~\cite{foti2021swapdisentangled}, where the latent representation of a mesh convolutional VAE is disentangled by curating the mini-batching procedure and introducing an additional loss. In particular, swapping shape attributes between the input meshes of every mini-batch, it is possible to know which of them share the same attribute and which share all the others. This knowledge is harnessed by a contrastive-like latent consistency loss that encourages subsets of latent variables from different meshes in the mini-batch to assume the same similarities and differences of the shapes created with the attribute swapping. This disentangles subsets of latent variables which become responsible for the generation of different body and head attributes. 
        We adopt the same network architecture, dataset, and attribute segmentation of SD-VAE. This choice is arbitrary and simplifies comparisons between the two methods, which differ only in their disentanglement technique. 
        
        Like VAEs, the research on GANs comes mostly from the imaging domain, where good levels of control over the generation process were recently made possible. Most of these models leverage segmentation maps~\cite{huang2021multimodal, lee2020maskgan, ling2021editgan}, additional attribute classifiers~\cite{he2019attgan, shoshan2021gan}, text prompts~\cite{radford21a}, or manipulate the latent codes and the parameter space of the pre-trained model to achieve the desired results~\cite{karras2021alias, harkonen2020ganspace, shen2020interfacegan, ling2021editgan}. 
        We argue that while the first two approaches require more inputs and supervision than our method, the last two offer less editing flexibility. In fact, describing the shape of human parts is a difficult task that would ultimately limit the diversity of the generated shapes, while the post-training manipulation may limit the exploration of some latent regions.
        Only a few methods explicitly seek disentanglement during training~\cite{alharbi2020disentangled, voynov2020unsupervised} like ours. However, \cite{alharbi2020disentangled} is specifically designed for grid-structured data, like images, and~\cite{voynov2020unsupervised} still requires a pre-trained GAN and two additional networks for disentanglement. In the 3D shapes domain, GAN disentanglement is still researched to control subject poses and expressions~\cite{chen2021intrinsic, olivier2021facetunegan} or object parts~\cite{li2021sp}. However, they suffer the same problems described for 3D VAEs: they have complex architectures and do not have control over the generation of local identity attributes.
        
    \subsection{Spectral Geometry.}
        Spectral mesh processing has played an essential role in shape indexing, sequencing, segmentation, parametrisation, correspondence, and compression~\cite{zhang2010spectral}. Spectral methods usually leverage the properties of the eigenstructures of operators such as the mesh Laplacian. Even though there is no unique definition for this linear operator, it can be classified either as geometric or combinatorial. Geometric Laplacians are a discretisation of the continuous Laplace-Beltrami operator~\cite{chavel1984eigenvalues} and, as their name suggests, they encode geometric information. Their eigenvalues are robust to changes in mesh connectivity and are often used as shape descriptors\cite{reuter2006laplace, gao2014compact}. Since they are isometry-invariant, they are used also in VAEs for identity and pose disentanglement~\cite{aumentado2019geometric, aumentado2021disentangling}. However, being geometry dependant, the Laplace-Beltrami operator and its eigendecomposition have to be precomputed for every mesh in the dataset. On the other hand, combinatorial Laplacians treat a mesh as a graph and are entirely defined by the mesh topology. For these operators, the eigenvectors can be considered as Fourier bases and the eigenprojections are equivalent to a Fourier transformation~\cite{shuman2013emerging} whose result is often used as a shape descriptor. If all shapes in a dataset share the same topology, the combinatorial Laplacian and its eigendecomposition need to be computed only once. For this reason, multiple graph and mesh convolutions~\cite{bruna2013spectral, defferrard2016convolutional} as well as some data augmentation technique~\cite{foti2020intraoperative} and smoothing losses~\cite{foti2021swapdisentangled} are based on combinatorial Laplacian formulations. 


\section{Method}
    The proposed method introduces a novel loss to improve latent disentanglement in generative models of 3D human shapes. After defining the adopted shape representation, we introduce our local eigenprojection loss, followed by the two generative models on which it was tested: a VAE and two flavours of GANs.
    
    \begin{figure}[t]
        \centering
        \includegraphics[width=\linewidth]{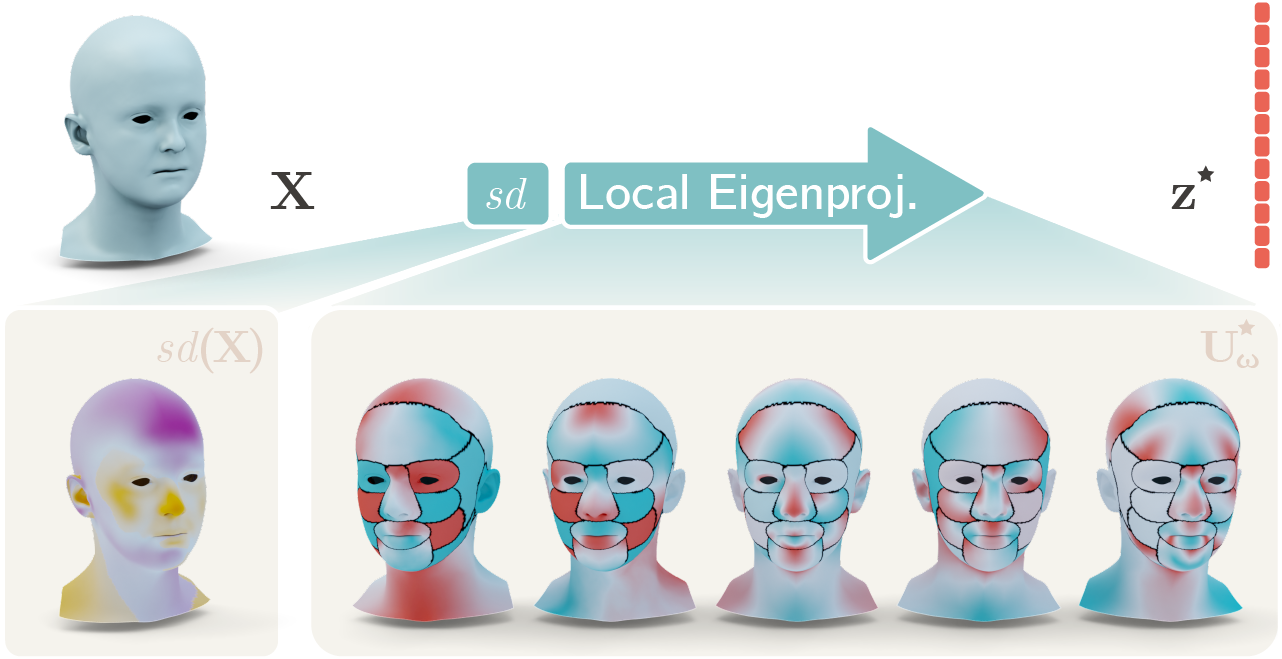}
        \caption{Schematic representation of the local eigenprojection, the operation at the core of our local eigenprojection loss. The signed distance between a given mesh $\mathbf{X}$ and a mean shape template is computed as $sd(\mathbf{X})$. $sd(\mathbf{X})$ is locally eigenprojected into a vector $\mathbf{z}^\star$ where each subset of variables is a spectral descriptor of a shape attribute. The projection is performed by matrix-multiplying the signed distance by $\mathbf{U}^\star_\omega$, the highest-variance eigenvectors of each shape attribute $\omega$. The heads in the bottom part of the figure represent one-dimensional vectors whose values are mapped with diverging colour maps on the mean shape head. On the heads corresponding to the columns of $\mathbf{U}^\star_\omega$, the black seams mark the different attributes 
        that we seek to control during the generation procedure.}
        \label{fig:eigproj}
    \end{figure}
    
    \subsection{Shape Representation.} 
        We represent 3D shapes as manifold triangle meshes with a fixed topology. By fixing the topology, all meshes $\mathcal{M} = \{\mathbf{X}, \boldsymbol{\mathcal{E}}, \boldsymbol{\mathcal{F}}\}$ share the same edges $\mathbf{\mathcal{E}} \in \mathbb{N}^{\varepsilon \times 2}$ and faces $\boldsymbol{\mathcal{F}} \in \mathbb{N}^{\Gamma \times 3}$. Therefore, they differ from one another only for the position of their vertices $\mathbf{X} \in \mathbb{R}^{N \times 3}$, which are assumed to be consistently aligned, scaled, and with point-wise correspondences across shapes. 
        
    \subsection{Local Eigenprojection Loss.}
        We define $F$ arbitrary attributes on a mesh template by manually colouring anatomical regions on its vertices. Thanks to the assumption of our shape representation, the segmentation of the template mesh can be consistently transferred to all the other meshes without manually segmenting them. Mesh vertices can be then grouped per-attribute such that $\mathbf{X} = \{ \mathbf{X}_\omega\}_{\omega = 1}^F$. 
        Seeking to train generative models capable of controlling the position of vertices corresponding to each shape attribute $\mathbf{X}_\omega$ through a predefined set of latent variables, we evenly split the latent representation $\mathbf{z}$ in $F$ subsets of size $\kappa$, such that $\mathbf{z} = \{ \mathbf{z}_\omega\}_{\omega = 1}^F$ and each $\mathbf{z}_\omega$ controls its corresponding $\mathbf{X}_\omega$. To establish and enforce a direct relationship between each $\mathbf{X}_\omega$ and $\mathbf{z}_\omega$ we rely on spectral geometry and compute low-dimensional local shape descriptors in the spectral domain. We start by computing the Kirchoff graph Laplacian corresponding to each shape attribute as: $\mathbf{K}_\omega = \mathbf{D}_\omega - \mathbf{A}_\omega$, where $\mathbf{A_\omega} \in \mathbb{N}^{N_\omega \times N_\omega}$ is the adjacency matrix of attribute $\omega$, $\mathbf{D}_\omega \in \mathbb{R}^{N_\omega \times N_\omega}$ its diagonal degree matrix, and $N_\omega$ the number of its vertices. Values on the diagonal of $\mathbf{D}_\omega$ are computed as $\text{D}_{aa} = \sum_b \text{A}_{ab}$. The Kirchoff Laplacian is a real symmetric positive semidefinite matrix that can be eigendecomposed as $\mathbf{K}_\omega = \mathbf{U}_\omega \boldsymbol{\Lambda}_\omega \mathbf{U}_\omega^T$. The columns of $\mathbf{U}_\omega \in \mathbb{R}^{N_\omega \times K}$ are a set of $K$ orthonormal eigenvectors known as the graph Fourier modes and can be used to transform any discrete function defined on the mesh vertices into the spectral domain. The signal most commonly transformed is the mesh geometry, which is the signal specifying the vertex coordinates. However, the local eigenprojection $\mathbf{\tilde{X}}_\omega = \mathbf{U}_\omega^T \mathbf{X}_\omega$ would result in a matrix of size $K \times 3$ containing the spectral representations of the $3$ spatial coordinates. Instead of flattening $\mathbf{\tilde{X}}_\omega$ to make it compatible with the shape of the latent representation, we define and project a one-dimensional signal: the signed distance between the vertices of a mesh and the per-vertex mean of the training set $\mathbf{M}$ (see Fig.~\ref{fig:eigproj}).
        We have:
        \begin{equation}
            \label{eq:signed_distance}
            sd(\mathbf{X}) = \boldsymbol{\gamma} \|\mathbf{X} - \mathbf{M} \|_2 \qquad \text{with} \quad \boldsymbol{\gamma} = \text{sign}(\big\langle \mathbf{X} - \mathbf{M}, \mathbf{N} \big\rangle),
        \end{equation}
        where $\langle \cdot, \cdot \rangle$ is the inner product, and $\mathbf{N}$ are the vertex normals referred to the mesh template with vertex positions $\mathbf{M}$.
        If X was standardised by subtracting $\mathbf{M}$ and dividing by the per-vertex standard deviation of the training set $\boldsymbol{\Sigma}$, being $\odot$ the Hadamard product, Eq.~\ref{eq:signed_distance} can be rewritten as: 
        \begin{equation}
            \label{eq:signed_distance_normalised}
            sd(\mathbf{X}) = \boldsymbol{\gamma} \|\mathbf{X} \odot \boldsymbol{\Sigma} \|_2 \qquad \text{with} \quad \boldsymbol{\gamma} = \text{sign}(\big\langle \mathbf{X} \odot \boldsymbol{\Sigma}, \mathbf{N} \big\rangle).
        \end{equation}
        We assume that not all eigenprojections are equally significant when representing shapes. Therefore, for each attribute $\omega$, we eigenproject all the local signed distances $sd(\mathbf{X}_\omega)$ computed over the training set, and identify the $\kappa$ (with $\kappa \ll K$) spectral components with the highest variance. While these spectral components are responsible for most shape variations, the small shape differences represented by other components can be easily learned by the neural-network-based generative model. After eigenprojecting the entire training set, we select the Fourier modes $\mathbf{U}_\omega^\star \in \mathbb{R}^{N_\omega \times \kappa}$ associated with the highest variance eigenprojections (Fig.~\ref{fig:eigproj}) and use them to compute the eigenprojection loss. During this preprocessing step we also compute the mean and standard deviation of the highest variance local eigenrpojections, which we denote by $\mathbf{m}_\omega^\star$ and $\mathbf{s}_\omega^\star$ respectively. We thus define the local eigenprojection loss as:
        \begin{equation}
            \label{eq:lep_loss}
            \mathcal{L}_\textit{LE}(\mathbf{X}, \mathbf{z}) = \frac{1}{F \kappa} \sum_{\omega=1}^{F} \Big\| \mathbf{z}_\omega - \frac{ (\mathbf{U}_\omega^\star)^T sd(\mathbf{X}_\omega)- \mathbf{m}_\omega^\star}{\mathbf{s}_\omega^\star} \Big\|_1
        \end{equation}
        Note that combinatorial Laplacian operators are determined exclusively by the mesh topology. Since the topology is fixed across the dataset, the Laplacians and their eigendecompositions can be computed only once. Therefore, the local eigenprojection can be quickly determined by matrix-multiplying signed distances by the precomputed $\mathbf{U}_\omega^\star$.
        Also, if the Laplace-Beltrami operator was used in place of the Kirchoff graph Laplacian, the eigendecomposition would need to be computed for every mesh. Not only this would significantly increase the training duration, but backpropagating through the eigendecomposition would be more complex as this would introduce numerical instabilities~\cite{wang2019backpropagation}.
        Alternatively, an approach similar to~\cite{marin2021spectral} should be followed.
        
        \begin{figure*}
                \centering
                \includegraphics[width=\linewidth]{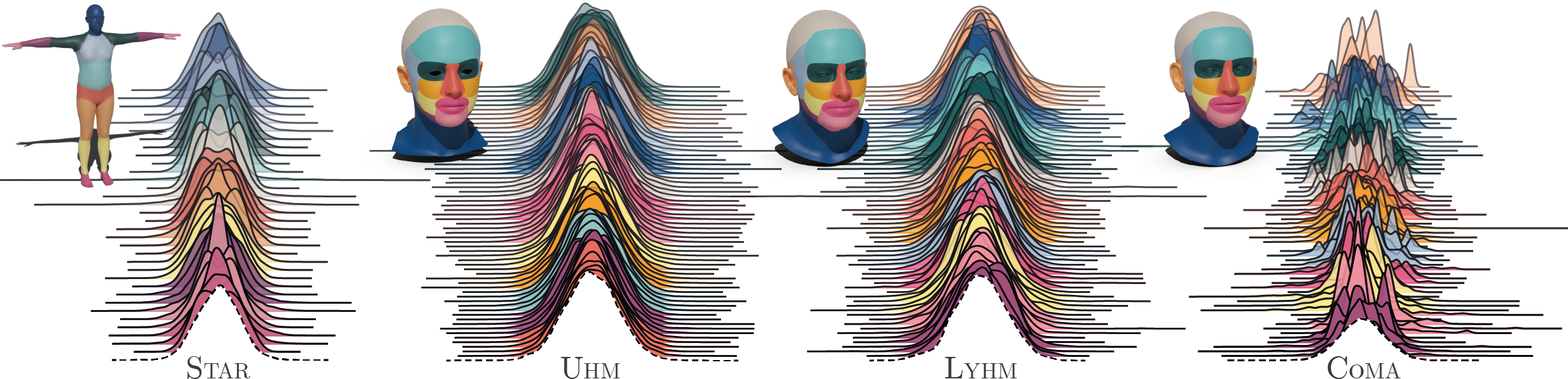}
                \caption{Local eigenprojection distributions. All training meshes are locally eigenprojected to observe the distributions of the elements in the resulting vectors. Distributions are colour-coded according to the shape attribute they are referred to. The segmentation of the shape attributes displayed next to the distributions is rendered on the mean shape templates of the corresponding dataset. The dashed distributions, which are obtained sampling a Gaussian, are reported for comparison.}
                \label{fig:eigproj_distrib}
        \end{figure*}  
        
        \subsection{Mesh Variational Autoencoder.}
            Like traditional VAEs~\cite{kingma2013auto}, our 3D-VAE is also built as a probabilistic encoder-decoder pair parameterised by two separate neural networks. The probabilistic encoder is defined as a variational distribution $q(\mathbf{z}|\mathbf{X})$ that approximates the intractable model posterior. It predicts the mean $\boldsymbol{\mu}$ and standard deviation $\boldsymbol{\sigma}$ of a Gaussian distribution over the possible $\mathbf{z}$ values from which $\mathbf{X}$ could have been generated. The probabilistic decoder $p(\mathbf{X}|\mathbf{z})$ describes the distribution of the decoded variable given the encoded one. During the generation process, a latent vector $\mathbf{z}$ is sampled from a Gaussian prior distribution $p(\mathbf{z}) = \mathcal{N}(\mathbf{z}; \mathbf{0}, \mathbf{I})$ and an output shape is generated by the probabilistic decoder. Since the decoder is used as a generative model, it is also referred to as generator. Following this convention, we define our architecture as a pair of non-linear functions $\{E, G\}$, where $E: \mathcal{X} \rightarrow \mathcal{Z}$ maps from the vertex embedding domain $\mathcal{X}$ to the latent distribution domain $\mathcal{Z}$, and $G: \mathcal{Z} \rightarrow \mathcal{X}$ vice versa. Since traditional convolutional operators are not compatible with the non-Euclidean nature of meshes, we build both networks as in~\cite{foti2021swapdisentangled}, using the simple yet efficient spiral convolutions~\cite{gong2019spiralnet++} and sparse matrix multiplications with transformation matrices obtained with quadric sampling~\cite{gong2019spiralnet++, ranjan2018generating} (see Supplementary Materials for more details).
            
            As in~\cite{foti2021swapdisentangled}, the 3D-VAE is trained minimising $\mathcal{L}_\textit{VAE} = \mathcal{L}_\textit{R}(\mathbf{X}, \mathbf{X}') + \alpha \mathcal{L}_\textit{L}(\mathbf{X}') + \beta \mathcal{L}_\textit{KL}(\boldsymbol{\mu}, \boldsymbol{\sigma})$. While $\alpha$ and $\beta$ are weighting constants, $\mathcal{L}_\textit{R}$ is the reconstruction loss, $\mathcal{L}_\textit{L}$ is a Laplacian regulariser, and $\mathcal{L}_\textit{KL}$ is a Kullback–Leibler (KL) divergence. In auto-encoder parlance, the reconstruction loss $\mathcal{L}_\textit{R}(\mathbf{X}, \mathbf{X}')=\frac{1}{N} \|\mathbf{X}' - \mathbf{X}\|_\text{F}^2$ encourages the output of the VAE to be as close as possible to its input by computing the squared Frobenius norm between $\mathbf{X}'=G(E(\mathbf{X}))$ and $\mathbf{X}$. The KL divergence can be considered as a regularisation term that pushes the variational distribution $q(\mathbf{z}|\mathbf{X})$ towards the prior distribution $p(\mathbf{z})$. Since both prior and posterior are assumed to be Gaussian $\mathcal{L}_\textit{KL}(\boldsymbol{\mu}, \boldsymbol{\sigma}) = \boldsymbol{\sigma}^2 + \boldsymbol{\mu}^2 - \text{log}(\boldsymbol{\sigma}^2) - 1$. The Laplacian loss $\mathcal{L}_\textit{L}(\mathbf{X}') = \frac{1}{N} \| \mathbf{TX}' \|_\text{F}^2$ is a smoothing term computed on the output vertices $\mathbf{X}'$ and based on the Tutte Laplacian $\mathbf{T} = \mathbf{D}^{-1}\mathbf{K} = \mathbf{I} - \mathbf{D}^{-1}\mathbf{A}$, where $\mathbf{A}$, $\mathbf{D}$, and $\mathbf{K}$ are the adjacency, diagonal degree, and Kirchoff Laplacian introduced in the previous paragraph and computed on the entire mesh rather than on shape attributes.
            
            Latent disentanglement is enforced by separately applying the local eigenprojection loss to the encoder and generator. We thus define the total loss as:
            \begin{equation}
                \label{eq:vae_disent_tot}
                \begin{split}
                    \mathcal{L} = \mathcal{L}_\textit{R}(\mathbf{X}, \mathbf{X}') + \alpha \mathcal{L}_\textit{L}(\mathbf{X}') + \beta \mathcal{L}_\textit{KL}(\boldsymbol{\mu}, \boldsymbol{\sigma}) + \\ + \eta_1 \mathcal{L}_\textit{LE}(\mathbf{X}, \boldsymbol{\mu}) + \eta_2 \mathcal{L}_\textit{LE}(\mathbf{X'}, \boldsymbol{\mu}),
                \end{split}
            \end{equation}
            where $\eta_1$ and $\eta_2$ are two scalar weights balancing the contributions of the two local eigenprojection losses. Note that $\mathcal{L}_\textit{LE}(\mathbf{X}, \boldsymbol{\mu})$ is backpropagated only through $E$. This term pushes the predicted $\boldsymbol{\mu}$ towards the standardised local eigenprojections of the input, while the KL divergence attempts to evenly distribute the encodings around the centre of the latent space.
            Similarly, $\mathcal{L}_\textit{LE}(\mathbf{X'}, \boldsymbol{\mu})$ is backpropagated only through $G$ and it enforces the output attributes to have an eigenprojection compatible with the predicted mean. 
            
        \subsection{Mesh Generative Adversarial Networks.}
            We propose two flavours of 3D Generative Adversarial Networks: one based on Least Squares GAN (LSGAN)~\cite{mao2017least} and one on Wasserstein GAN (WGAN)~\cite{arjovsky2017wasserstein}. Like VAEs, GANs also rely on a pair of neural networks: a generator-discriminator pair $\{G, D\}$ in LSGAN and a generator-critic $\{G, C\}$ pair in WGAN. The architecture of the generators is the same as the one adopted in the generator of the 3D-VAE. The architectures of $D$ and $C$ are similar to $E$, but 
            with minor differences in the last layers (see Supplementary Materials). 
            Nevertheless, all networks are built with the same mesh operators of our 3D-VAE and~\cite{foti2021swapdisentangled, gong2019spiralnet++}. 
            
            In the LSGAN implementation, $G$ samples an input latent representation from a Gaussian distribution $p(\mathbf{z}) = \mathcal{N}(\mathbf{z}; \mathbf{0}, \mathbf{I})$ and maps it to the shape space as $G(\mathbf{z}) = \mathbf{X}'$. While it tries to learn a distribution over generated shapes, the discriminator operates as a classifier trying to distinguish generated shapes $\mathbf{X}'$ from real shapes $\mathbf{X}$. Using a binary coding scheme for the labels of real and generated samples, we can write the losses of $G$ and $D$ respectively as $\mathcal{L}^\textit{G}_\textit{LSGAN} = \frac{1}{2} \mathbb{E}_{\mathbf{z} \sim p(\mathbf{z})} [(D(G(\mathbf{z})) - 1)^2]$ and $\mathcal{L}^\textit{D}_\textit{LSGAN} = \frac{1}{2} \mathbb{E}_{\mathbf{X} \sim p(\mathbf{X})} [(D(\mathbf{X}) - 1)^2] + \frac{1}{2} \mathbb{E}_{\mathbf{z} \sim p(\mathbf{z})} [D(G(\mathbf{z}))^2]$. 
            We also add the Laplacian regularisation term $\mathcal{L}_\textit{L}(\mathbf{X}')$ to smooth the generated outputs. When seeking disentanglement, we train the discriminator by minimising $\mathcal{L}^\textit{D}_\textit{LSGAN}$ and the generator by minimising the following:
            \begin{equation}
                \label{eq:lsgan_gen_disent_tot}
                \mathcal{L}^G_\textit{LS} = \mathcal{L}^\textit{G}_\textit{LSGAN} + \alpha \mathcal{L}_\textit{L}(\mathbf{X}') + \eta \mathcal{L}_\textit{LE}(\mathbf{X'}, \mathbf{z}).
            \end{equation}
            In WGAN, $G$ still tries to learn a distribution over generated shapes, but its critic network $C$, instead of classifying real and generated shapes, learns a Wasserstein distance and outputs scalar scores that can be interpreted as measures of realism for the shapes it processes. The WGAN losses for $G$ and $C$ are $\mathcal{L}^\textit{G}_\textit{WGAN} = - \mathbb{E}_{\mathbf{z} \sim p(\mathbf{z})} [D(G(\mathbf{z})]$ and $\mathcal{L}^\textit{C}_\textit{WGAN} = \mathbb{E}_{\mathbf{z} \sim p(\mathbf{z})} [D(G(\mathbf{z}))] - \mathbb{E}_{\mathbf{X} \sim p(\mathbf{X})} [D(\mathbf{X})]$ respectively. Similarly to the LSGAN implementation, when enforcing disentanglement, the critic is trained minimising $\mathcal{L}^\textit{C}_\textit{WGAN}$, while the generator minimising:
            \begin{equation}
                \label{eq:wgan_gen_disent_tot}
                \mathcal{L}^G_\textit{W} = \mathcal{L}^\textit{G}_\textit{WGAN} + \alpha \mathcal{L}_\textit{L}(\mathbf{X}') + \eta \mathcal{L}_\textit{LE}(\mathbf{X'}, \mathbf{z}).
            \end{equation}
            Note that to make $C$ a 1-Lipschitz function, and thus satisfies the Wasserstein distance computation requirements, $C$ weights are clipped to the range $[-c, c]$.

\section{Experiments}
    \label{sec:experiments}
    \begin{table*}
        \centering
        \caption{Quantitative comparison between our model and other state-of-the-art methods. All methods were trained on \textsc{Uhm}~\cite{ploumpis2019combining}. Diversity, JSD, MMD, COV, and 1-NNA evaluate the generation capabilities of the models, while VP evaluates latent disentanglement.  The different metrics are computed as detailed in Sec.~\ref{sec:experiments_comparisons}. Note that the training time does not consider the initialisation time.
        }
        \label{tab:recon-div}
        \begin{tabular}{l c c c c c c c c}
            \toprule
                \multirow{2}{*}{Method} & \multirow{2}{*}{\parbox[c]{.1\linewidth}{\centering Mean Rec. ($\downarrow$)}} & \multirow{2}{*}{\parbox[c]{.08\linewidth}{\centering Diversity ($\uparrow$)}} & \multirow{2}{*}{\parbox[c]{.08\linewidth}{\centering JSD \\ ($\downarrow$)}} & \multirow{2}{*}{\parbox[c]{.08\linewidth}{\centering MMD \\ ($\downarrow$)}} & \multirow{2}{*}{\parbox[c]{.08\linewidth}{\centering COV \\ (\%, $\uparrow$)}} & \multirow{2}{*}{\parbox[c]{.08\linewidth}{\centering 1-NNA \\ ($\Delta$\%, $\downarrow$)}} &
                \multirow{2}{*}{\parbox[c]{.08\linewidth}{\centering VP \\ (\%, $\uparrow$)}} & \multirow{2}{*}{\parbox[c]{.1\linewidth}{\centering Training Time ($\downarrow$)}} \\ \\
            \midrule
                VAE                    & $\mathbf{0.61}$ & $4.23$      & $4.89$      & $1.53$      & $65.49$      & $1.17$         & $63.73$    & $\mathbf{1\textbf{h:}46\textbf{m}}$ \\ 
                LSGAN                  & ---         & $6.12$      & $\mathbf{1.14}$ & $1.65$      & $43.41$      & $22.04$        & $46.83$    & $2$h:$23$m \\
                WGAN                   & ---         & $4.04$      & $22.75$     & $1.36$      & $57.94$      & $23.98$        & $71.07$    & $2$h:$22$m \\
                DIP-VAE-I              & $4.65$      & $4.74$      & $5.32$      & $\mathbf{1.24}$ & $55.57$      & $4.31$         & $35.60$    & $1$h:$48$m \\ 
                SD-VAE                 & $0.73$      & $4.23$      & $4.30$      & $1.56$      & $\mathbf{65.67}$ & $\mathbf{0.50}$    & $79.75$    & $7$h:$21$m \\ 
            \midrule
                LED-VAE                & $1.46$      & $5.30$      & $2.27$      & $1.73$      & $49.83$      & $15.80$        & $\mathbf{80.75}$ & $2$h:$53$m \\ 
                LED-LSGAN              & ---         & $\mathbf{6.38}$ & $2.09$      & $2.03$      & $43.41$      & $17.23$        & $79.75$      & $2$h:$28$m \\
                LED-WGAN               & ---         & $5.77$      & $2.55$      & $1.81$      & $47.47$      & $14.95$        & $74.11$      & $2$h:$28$m \\
            \bottomrule
        \end{tabular}
    \end{table*}

    \subsection{Datasets.}
        Since our main objective is to train a generative model capable of generating different identities, we require datasets containing a sufficient number of subjects in a neutral expression (pose). Most open source datasets for 3D shapes of faces, heads, bodies, or animals (e.g. \textsc{Mpi}-Dyna~\cite{pons2015dyna}, \textsc{Smpl}~\cite{loper2015smpl}, \textsc{Surreal}~\cite{varol2017learning}, Co\textsc{ma}~\cite{ranjan2018generating}, \textsc{Smal}~\cite{zuffi20173d}, etc.) focus on capturing different expressions or poses and are not suitable for identity disentanglement. For comparison, we rely on the $10,000$ meshes --with neutral expression and pose-- generated in~\cite{foti2021swapdisentangled} using two linear models that were built using a large number of subjects: \textsc{Uhm}~\cite{ploumpis2019combining} and \textsc{Star}~\cite{osman2020star} (Sec.~\ref{sec:related_work_gen_models}). We also use the same data split with $90\%$ of the data for training, $5\%$ for validation, and $5\%$ for testing. Since these data are generated from PCA-based models, we also train our models on real data from the \textsc{Lyhm} dataset~\cite{dai2020statistical} registered on the \textsc{Flame}~\cite{li2017learning} template. In addition, even though it is beyond the scope of this work, we attempt to achieve disentanglement through local eigenprojection also on Co\textsc{ma}~\cite{ranjan2018generating}, a dataset mostly known for its wide variety of expressions. All models and datasets are released for non-commercial scientific research purposes.
        
    \subsection{Local Eigenprojection Distributions.}
        We observe that the eigenprojections are normally distributed for datasets with neutral poses or expressions (Fig.~\ref{fig:eigproj_distrib}). By standardising the eigenprojections in Eq.~\ref{eq:lep_loss} we ensure their mean and standard deviation to be $0$ and $1$ respectively. Since we enforce a direct relation between the local eigenprojections and the latent representations, this is a desirable property that allows us to generate meaningful shapes by sampling latent vector from a normal distribution.
        In order to explain why this property holds for datasets with neutral poses and expressions, we need to hypothesise that shapes follow a Gaussian distribution. This is a reasonable hypothesis for datasets generated from PCA-based models, such as those obtained from \textsc{Uhm} and \textsc{Star}, because vertex positions are computed as linear combinations of generative coefficients sampled from a Gaussian. However, following the maximum entropy explanation~\cite{lyon2014normal}, it is also reasonable to assume that shapes in dataset obtained capturing real people (like \textsc{Lyhm}), are normally distributed. 
        \cite{lyon2014normal} argues that although the Central Limit Theorem is the standard explanation of why many things are normally distributed, the conditions to apply the theorem are usually not met or they cannot be verified. We assume that, like people's height, also body and head shapes are largely determined by genetics and partially by environment and epigenetic effects. The selection pressure determines an ideal shape with some variability to hedge against fluctuating circumstances in the environment. This amounts to fixing the mean, and an upper bound on the variance. Apart from that, the population will naturally tend to a state of maximal disorder (i.e., maximum entropy). Therefore, according to the maximum entropy explanation, human shapes are normally distributed because the distribution maximising entropy subject to those constraints is a normal distribution.
        If the shapes are normally distributed, we can consider also vertex positions consistently sampled on the shape surfaces to follow each a different Gaussian distribution centred at the corresponding vertex coordinates on the mean shape. Considering that the signed distance and the local eigenprojection are both linear operations, they preserve normality, and for this reason also the local eigenprojections are normal. Note that expressions are subject-specific deformations with a highly non-linear behaviour~\cite{chandran2020semantic}. There is no guarantee that these transformations preserve the normality of the shape distribution. Therefore, datasets containing expressions, such as Co\textsc{ma}, may not satisfy  the normality assumption. In fact, we observe that the standardised eigenprojections have more complex distributions which appear to be mixture of Gaussians (see Fig.~\ref{fig:eigproj_distrib}). Intuitively, each Gaussian in the mixtures could be related to a different subset of expressions. 
        
        \begin{figure*}
            \centering
            \includegraphics[width=\textwidth]{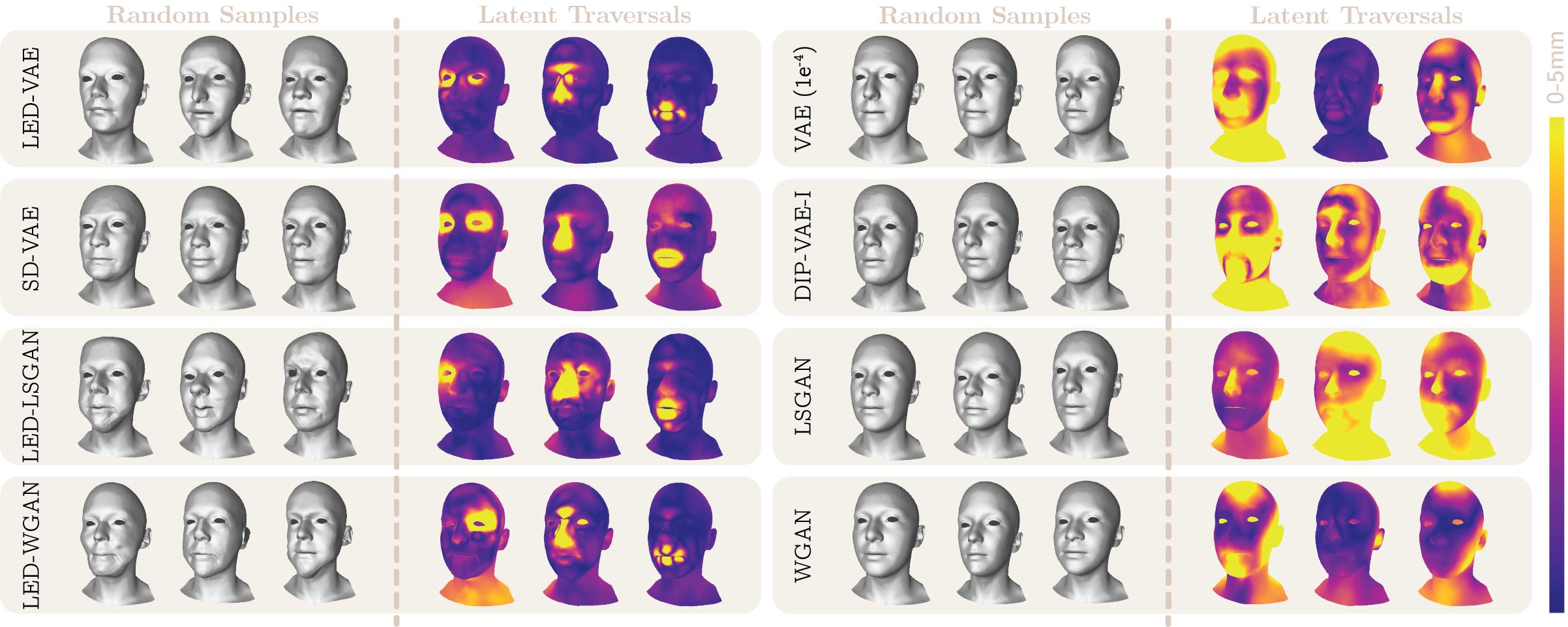}
            \caption{Random samples and vertex-wise distances showing the effects of traversing three randomly selected latent variables (see Supplementary Material to observe the effects for all the latent variables).}
            \label{fig:rnd_z_comp}
        \end{figure*}
        
    \subsection{Comparison with Other Methods.}
        \label{sec:experiments_comparisons}
        We compare our local eigenprojection disentangled (LED) methods against their vanilla implementations and against the only state-of-the-art method providing control over the generation of local shape attributes: the swap disentangled VAE (SD-VAE) proposed in~\cite{foti2021swapdisentangled}. The authors compared their SD-VAE with other VAEs for latent disentanglement. Among their implementation of DIP-VAE-I, DIP-VAE-II, and FactorVAE, the first one appeared to be the best performing. Therefore, we report results for DIP-VAE-I. 
        For a fair comparison, all methods were trained on the same dataset (\textsc{Uhm}) using the same batch size and the same number of epochs. In addition, they share the same architecture 
        with minor modifications for the GAN implementations (see Supplementary Materials). The SD-VAE implementation, as well as the evaluation code and the benchmark methods, are made publicly available at \href{https://github.com/simofoti/3DVAE-SwapDisentangled}{github.com/simofoti/3DVAE-SwapDisentangled}. All models were trained on a single Nvidia Quadro P5000, which was used for approximately 18 GPU days in order to complete all the experiments.
        
        The reconstruction errors reported in Tab.~\ref{tab:recon-div} are computed as the mean per-vertex L2 distance between input and output vertex positions. This metric is computed on the test set and applies only to VAEs. 
        We report the generation capabilities of all models in terms of diversity, JSD, MMD, COV, and 1-NNA. The diversity is computed as the average of the mean per-vertex distances among pairs of randomly generated meshes. The Jensen-Shannon Divergence (JSD)~\cite{achlioptas2018learning} evaluates the KL distances between the marginal point distributions of real and generated shapes. The coverage (COV)~\cite{achlioptas2018learning} measures the fraction of meshes matched to at least one mesh from the reference set. The minimum matching distance (MMD)~\cite{achlioptas2018learning} complements the coverage by averaging the distance between each mesh in the test set and its nearest neighbour among the generated ones. The 1-nearest neighbour accuracy (1-NNA) is a measure of similarity between shape distributions that evaluates the leave-one-out accuracy of a 1-NN classifier. In its original formulation~\cite{yang2019pointflow}, it expects values converging to $50\%$. However, following~\cite{foti2021swapdisentangled}, in Tab.~\ref{tab:recon-div} we report absolute differences between the original score and the $50\%$ target value. All the generation capability metrics can be computed either with the Chamfer or the Earth Mover distance. Since we did not observe significant discrepancies between the metrics computed with these two distances, we arbitrarily report results obtained with the Chamfer distance.

        \begin{figure*}
            \centering
            \includegraphics[width=\textwidth]{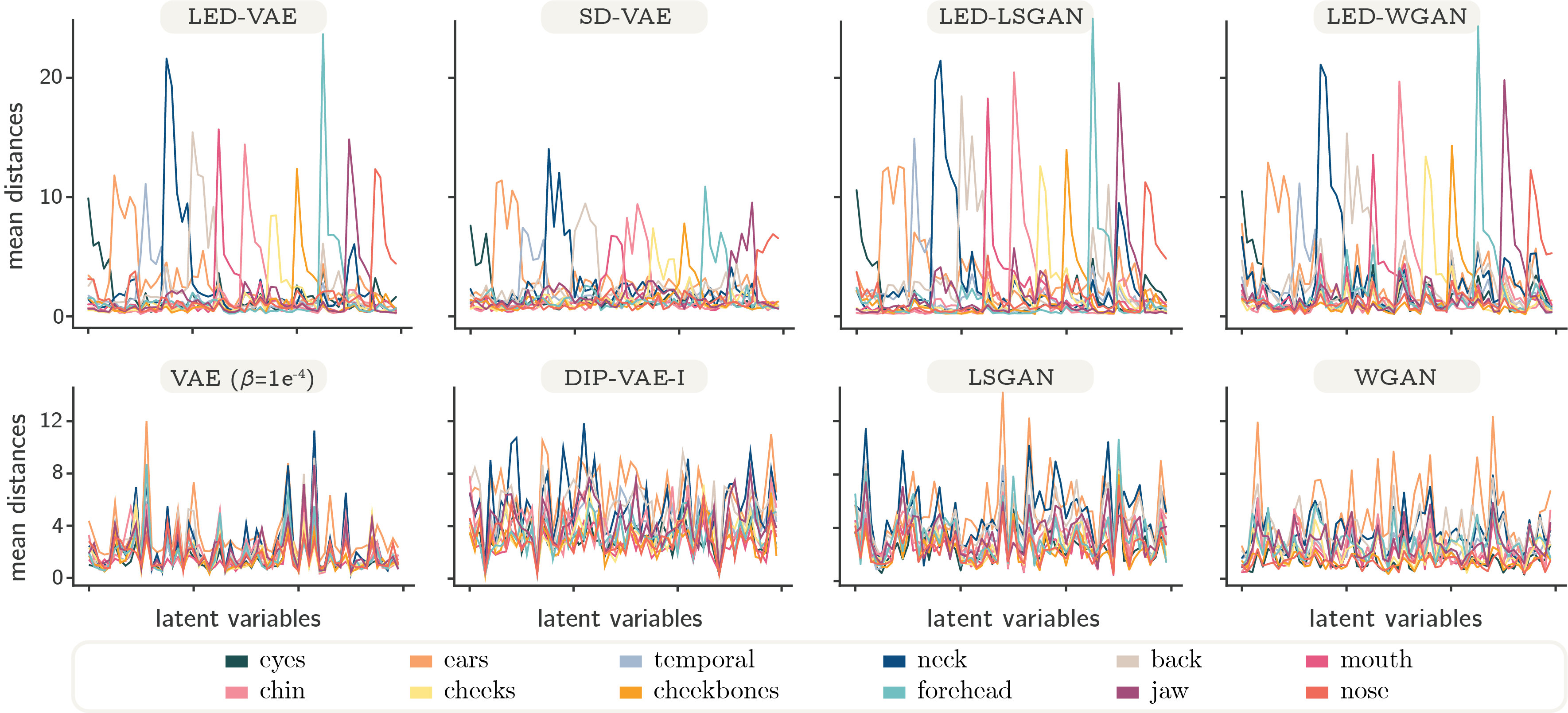}
            \caption{Effects of traversing each latent variable across different mesh attributes. For each latent variable (abscissas) we represent the per-attribute mean distances computed after traversing the latent variable from its minimum to its maximum value. 
            For each latent variable, we expect a high mean distance in one single attribute and low values for all the others.}
            \label{fig:z_compared_cond}
        \end{figure*}
        
        Observing Tab.~\ref{tab:recon-div} we notice that none of the models is consistently outperforming the others. GANs generally report better diversity scores than VAEs, but they are worse in terms of coverage and 1-NNA. GANs were also more difficult to train and were prone to mode collapse. On the other hand, VAEs appeared stable and required significantly less hyperparameter tuning. The scores of our LED models were comparable with other methods, thus showing that our loss does not negatively affect the generation capabilities. 
        However, LED models are consistently outperformed in terms of MMD, COV, and 1-NNA. These metrics evaluate the quality of generated samples by comparing them to a reference set. Since comparisons are performed on the entire output shapes, we hypothesise that a shape with local identity attributes resembling each a different subject from the test set is more penalised than a shape whose attributes are plausibly obtained from a single subject. Note also that MMD, COV, and 1-NNA appear to be inversely proportional to the diversity, suggesting that more diverse generated shapes are also less similar to shapes in the test set. LED-models report higher diversity because attributes can be independently generated. This negatively affects MMD, COV, and 1-NNA, but the randomly generated shapes are still plausible subjects (see Fig.~\ref{fig:rnd_z_comp} and Supplementary Materials). 
        Interestingly, SD-VAE appears to be still capable of generating shapes with attributes resembling the same subject from the test set, but at the expense of diversity and latent disengagement (see Sec.~\ref{sec:eval_disent}).
        
        LED-LSGAN and LED-WGAN train almost as quickly as the vanilla LSGAN and WGAN. Training LED-VAE takes approximately one hour more than its vanilla counterpart because the local eigenprojection loss is separately backpropagated through the encoder. However, since latent disentanglement is achieved without swapping shape attributes during mini-batching, the training time of LED-VAE is reduced by $61\%$ with respect to SD-VAE. Note that the additional initialisation overhead of LED models ($3.72$ minutes) is negligible when compared to the significant training time reduction over SD-VAE, which is the only model capable of achieving a satisfactory amount of latent disentanglement. 
        
        If we then qualitatively evaluate the random samples in Fig.~\ref{fig:rnd_z_comp}, we see that the quality of the meshes generated by LED-LSGAN and LED-WGAN is slightly worse than those from LED-VAE.
        We attribute this behaviour to the --usually undesired-- smoothness typically introduced by 3D VAE models. In this case, the VAE model itself acts as a regulariser that prevents the shape artefacts introduced by the local eigenprojection disentanglement. In addition, traversing the latent variables, we find that mesh defects tend to appear when latent variables approach values near $\pm 3$ (see supplementary material video). This might be a consequence of the reduced number of training data with local eigenprojections with these values (see Fig.~\ref{fig:eigproj_distrib}). Nonetheless, the problem can be easily mitigated with the truncation trick, thus sampling latent vectors from a Gaussian with standard deviation slightly smaller than one. 
        
        \begin{figure*}
            \centering
            \includegraphics[width=\textwidth]{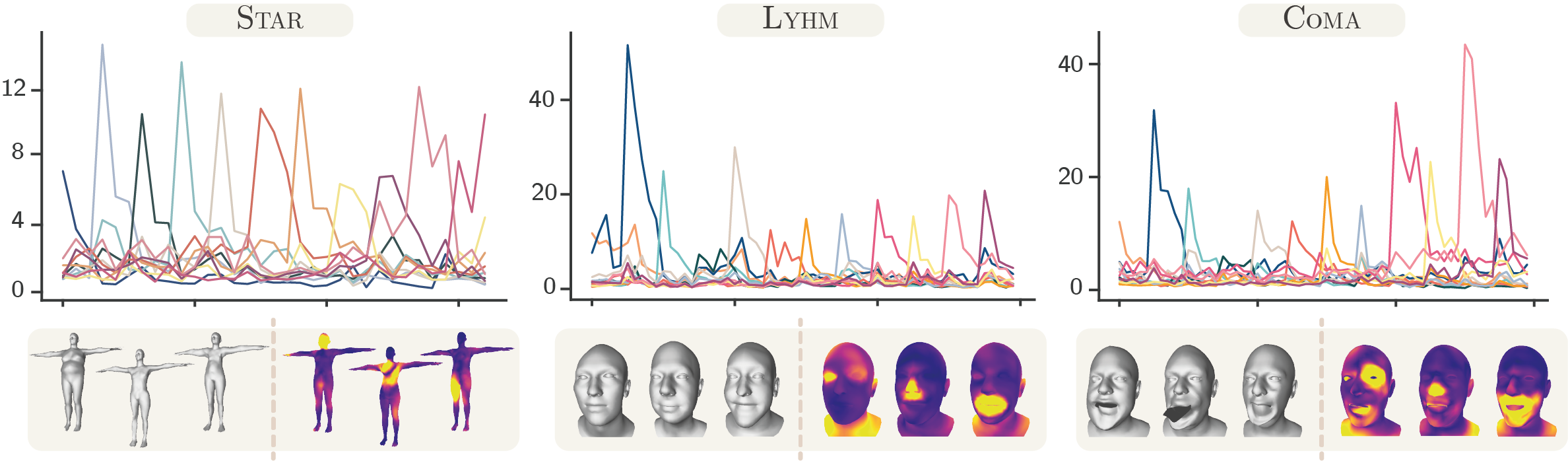}
            \caption{Results of LED-VAE on other datasets. For each dataset are displayed the effects of traversing latent variables (\textsc{Uhm} is reported in Fig.~\ref{fig:z_compared_cond}), three random samples and three vertex-wise distances highlighting the effects of traversing three latent variables (\textsc{Uhm} is reported Fig.~\ref{fig:rnd_z_comp}). Mean distances are plotted following the colour coding depicted in Fig.~\ref{fig:eigproj_distrib}.}
            \label{fig:other-datasets}
        \end{figure*}
        
    \subsection{Evaluation of Latent Disentanglement.}
        \label{sec:eval_disent}
        Latent disentanglement can be quantitatively evaluated on datasets with labelled data. However, such labels are not available for the disentanglement of shape attributes and traditional metrics such as Z-Diff~\cite{higgins2016beta}, SAP~\cite{kumar2017variational}, and Factor~\cite{kim2018disentangling} scores cannot be used. Since the Variation Predictability (VP) disentanglement metric does not require labelled data and it has shown good correlation with the Factor score~\cite{zhu2020learning}, we rely on this metric to quantify disentanglement across different models (see Tab.~\ref{tab:recon-div}). The VP metric averages the test accuracies across multiple few-shot trainings of a classifier. The classifier takes as input the difference between two shapes generated from two latent vectors differing in only one dimension and predicts the varied latent dimension. We implement the classifier network with the same architecture of our encoders, discriminators, and critiques. The network was trained for $5$ epochs with a learning rate of $1e^{-4}$. As in~\cite{zhu2020learning}, we set $\eta_{VP}=0.1$, $N_{VP}=10,000$ and $S_{VP}=3$.
        
        In addition, we qualitatively evaluate disentanglement as in~\cite{foti2021swapdisentangled} by observing the effects of traversing latent variables (Fig.~\ref{fig:teaser}, \textit{left}). 
        For each latent variable, we compute the per-vertex Euclidean distances between two meshes. After setting all latent variables to their mean value ($0$), the first mesh is generated setting a single latent to its minimum ($-3$) and the second mesh setting the same variable to its maximum ($+3$). The Euclidean distances can be either rendered on the mesh surface using colours proportional to the distances (Latent Traversals in Fig.~\ref{fig:rnd_z_comp} and Fig.~\ref{fig:other-datasets}), or plotted as their per-attribute average distance (Fig.~\ref{fig:z_compared_cond} and Fig.~\ref{fig:other-datasets}). 
        When plotted, the average distances isolated to each attribute 
        provide an intuitive way to assess disentanglement: good disentanglement is achieved when the traversal of a single variable determines high mean distances for one attribute and low mean distances for all the others.
        Observing Fig.~\ref{fig:rnd_z_comp} and Fig.~\ref{fig:z_compared_cond}, it is clear that the only state-of-the-art method providing control over local shape attributes is SD-VAE. Since the eigenvectors used in the local eigenprojection loss are orthogonal, we improve disentanglement over SD-VAE. In fact, traversing latent variables of LED models determines finer changes within each attribute in the generated shapes. For instance, this can be appreciated by observing the eyes of the latent traversals in Fig.~\ref{fig:rnd_z_comp}, where left and right eyes are controlled by different variables in LED-VAE, while by the same one in SD-VAE (more examples are depicted in the Supplementary Materials).  We also notice that the magnitude of the mean distances reported in Fig.~\ref{fig:z_compared_cond} for our LED models is bigger than SD-VAE within attributes and comparable outside. This shows superior disentanglement and allows our models to generate shapes with more diverse attributes than SD-VAE. Our model exhibits good disentanglement performances also on other datasets (Fig.~\ref{fig:other-datasets}).
    
    \subsection{Direct Manipulation}
        \label{a-subsec:manipulation}
\begin{figure}[b]
            \centering
            \includegraphics[width=\linewidth]{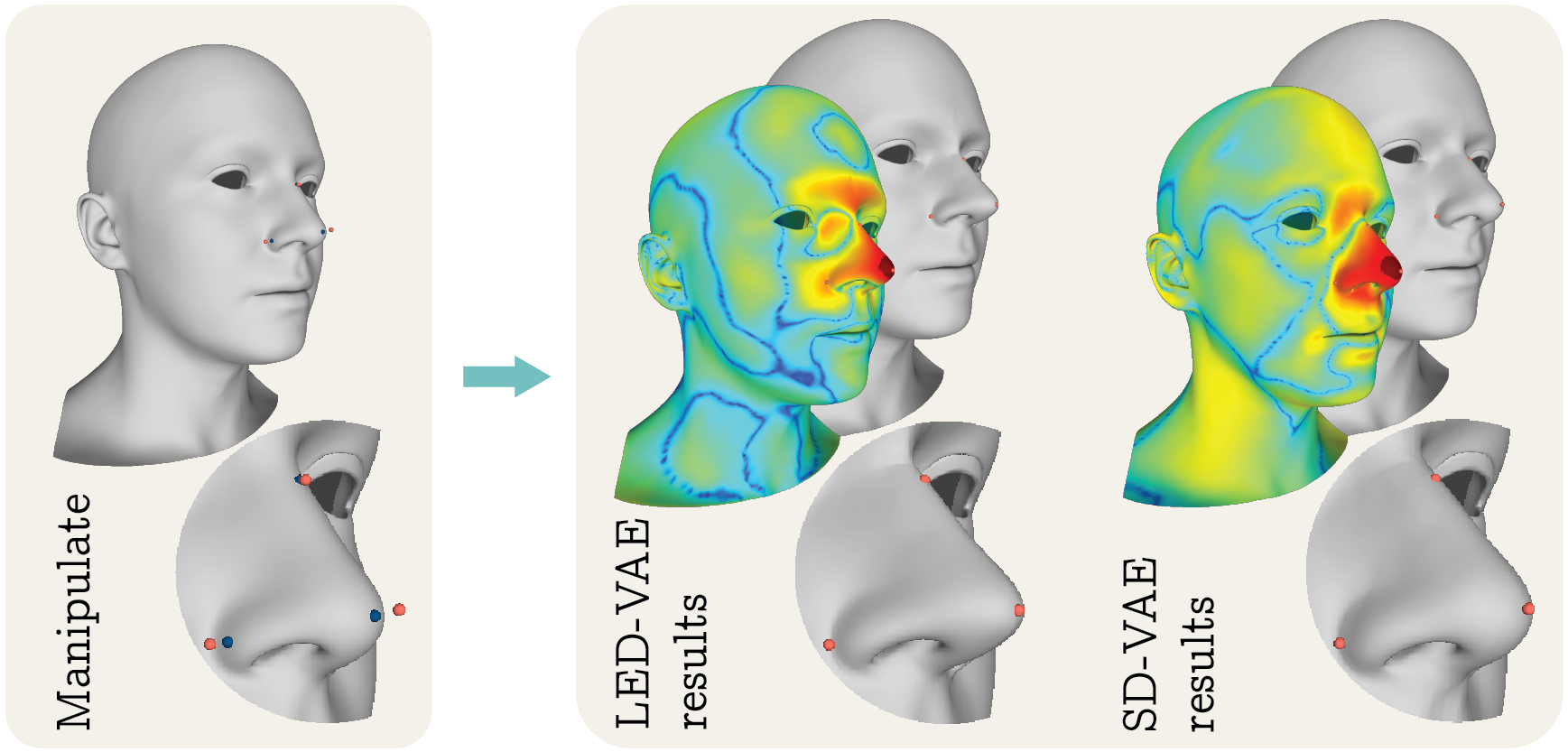}
            \caption{Direct manipulation. \textit{Left}: the user manually selects an arbitrary number of vertices (blue) and specifies their desired position (red). \textit{Right}: results of the direct manipulation optimisation for LED-VAE and SD-VAE. For each method, the output shape, a close-up of the manipulated attribute, and the rendering of the per-vertex distances between the initial and manipulated shapes are reported. The colour-map used to represent vertex distances is blue where distances are zero and red where they reach their maximum value. }
            \label{a-fig:direct-manip}
        \end{figure}

        Like SD-VAE, also LED-VAE can be used for the direct manipulation of the generated shapes. As in~\cite{foti2021swapdisentangled}, the direct manipulation is performed by manually selecting $\Upsilon$ vertices on the mesh surface ($ S \circ \mathbf{X}' = S \circ G(\mathbf{z}) \in \mathbb{R}^{\Upsilon \times 3}$) and by providing their desired location ($\mathbf{Y} \in \mathbb{R}^{\Upsilon \times 3}$). Then, $\min_{\mathbf{z}_\omega} \| S \circ G(\mathbf{z}) - \mathbf{Y}\|_2^2$ is optimised with the \textsc{Adam} optimiser for $50$ iterations while maintaining a fixed learning rate of $lr=0.1$. Note that the optimisation is performed only on the subset of latent variables $\mathbf{z}_\omega$ controlling the local attribute corresponding to the selected vertices. If vertices from different attributes are selected, multiple optimisations are performed. As it can be observed in Fig.~\ref{a-fig:direct-manip}, LED-VAE is able to perform the direct manipulations causing less shape changes than SD-VAE in areas that should remain unchanged.  
        
\section{Conclusion}
    \label{sec:conclusion}  
    We introduced a new approach to train generative models with a more disentangled, interpretable and structured latent representation that significantly reduces the computational burden required by SD-VAE. By establishing a correspondence between local eigenprojections and latent variables, generative models can better control the creation and modification of local identity attributes of human shapes (See Fig.~\ref{fig:teaser}). Like the majority of state-of-the-art methods, the main limitation of our model is the assumption on the training data, which need to be consistently aligned, in dense point correspondence, and with a fixed topology. Even though this is surely a limitation, as we mentioned in Sec.~\ref{sec:intro}, this assumption can simplify the generation of other digital human's properties. Among the different LED models we proposed, we consider LED-VAE to be the most promising. This model is simpler to train, requires less hyperparameter tuning, and generates higher-quality meshes. We trained and tested this model also on other datasets, where it showed equivalent performances. Datasets with expressions have complex local eigenprojection distributions (Fig.~\ref{fig:eigproj_distrib}) which are more difficult to learn. In fact, random samples generated by LED-VAE trained on Co\textsc{ma} present mesh defects localised especially in areas where changes in expression introduce significant shape differences characterised by a highly non-linear behaviour (e.g. the mouth region). Controlling the generation of different expressions was beyond the scope of this work and we aim at addressing the issue as future work.  
    We proved that our loss can be easily used with both GANs and VAEs. Being efficient to compute and not requiring modifications to the mini-batching procedure (like SD-VAE), it could be leveraged also in more complex architectures for 3D reconstruction or pose and expression disentanglement. In the LED-VAE the local eigenprojection loss is computed also on the encoder (see how this improves disentanglement in the ablation study provided with the supplementary materials). Having an encoder capable of providing a disentangled representation for different attributes could greatly benefit shape-analysis research in plastic surgery~\cite{o2022convolutional} and in genetic applications~\cite{claes2018genome}.
    Therefore, we believe that our method has the potential to benefit not only experienced digital artists but also democratise the creation of realistic avatars for the metaverse and find new applications in shape analysis. Since the generation of geometric shapes is only the first step towards the data-driven generation of realistic digital humans, as future work, we will research more interpretable generative processes for expressions, poses, textures, materials, high-frequency details, and hair.
    
\section*{Acknowledgement}
    This research was funded in whole, or in part, by the Wellcome Trust [203145Z/16/Z]. For the purpose of Open Access, the author has applied a CC BY public copyright licence to any Author Accepted Manuscript version arising from this submission.


\printbibliography                

@String{Computing = "Computing" }

@String{Computer = "{IEEE} Computer" }

@String{Academic = "Academic Press" }

@String{Springer = "Springer-Verlag" }

@inproceedings{foti2021swapdisentangled,
  title={3D Shape Variational Autoencoder Latent Disentanglement via Mini-Batch Feature Swapping for Bodies and Faces},
  author={Foti, Simone and Koo, Bongjin and Stoyanov, Danail and Clarkson, Matthew J},
  booktitle={Proceedings of the IEEE/CVF Conference on Computer Vision and Pattern Recognition},
  pages={18730--18739},
  year={2022},
  publisher={IEEE},
  address={New Orleans, Louisiana, USA}
}

@inproceedings{foti2020intraoperative,
  author = {Foti, Simone and Koo, Bongjin and Dowrick, Thomas and Ramalhinho, Jo\~{a}o and Allam, Moustafa and Davidson, Brian and Stoyanov, Danail and Clarkson, Matthew J.},
  title = {Intraoperative Liver Surface Completion with Graph Convolutional VAE},
  year = {2020},
  isbn = {978-3-030-60364-9},
  publisher = {Springer-Verlag},
  address = {Berlin, Heidelberg},
  booktitle = {Uncertainty for Safe Utilization of Machine Learning in Medical Imaging, and Graphs in Biomedical Image Analysis},
  pages = {198–207},
  numpages = {10},
  location = {Lima, Peru}
}

@article{zhang2010spectral,
  author={Zhang, Hao and Van Kaick, Oliver and Dyer, Ramsay},
  title = {Spectral Mesh Processing},
  journal = {Computer Graphics Forum},
  volume = {29},
  number = {6},
  pages = {1865-1894},
  year = {2010}
}

@article{gruber2020interactive,
  author = {Gruber, A. and Fratarcangeli, M. and Zoss, G. and Cattaneo, R. and Beeler, T. and Gross, M. and Bradley, D.},
  title = {Interactive Sculpting of Digital Faces Using an Anatomical Modeling Paradigm},
  journal = {Computer Graphics Forum},
  booktitle={Computer Graphics Forum},
  organization={Wiley Online Library},
  volume = {39},
  number = {5},
  pages = {93-102},
  year = {2020}
}

@article{bengio2013representation,
  title={Representation learning: A review and new perspectives},
  author={Bengio, Yoshua and Courville, Aaron and Vincent, Pascal},
  journal={IEEE transactions on pattern analysis and machine intelligence},
  volume={35},
  number={8},
  pages={1798--1828},
  year={2013},
  publisher={IEEE}
}

@misc{higgins2016beta,
  title={beta-{VAE}: Learning Basic Visual Concepts with a Constrained Variational Framework},
  author={Irina Higgins and Loic Matthey and Arka Pal and Christopher Burgess and Xavier Glorot and Matthew Botvinick and Shakir Mohamed and Alexander Lerchner},
  booktitle={5th International Conference on Learning Representations},
  year={2017},
  address={Toulon, France},
  publisher={OpenReview.net}
}

@inproceedings{kim2018disentangling,
  title={Disentangling by factorising},
  author={Kim, Hyunjik and Mnih, Andriy},
  booktitle={International Conference on Machine Learning},
  pages={2649--2658},
  year={2018},
  organization={PMLR},
  publisher={PMLR},
  address={Stockhol, Sweden}
}

@inproceedings{ding2020guided,
  title={Guided variational autoencoder for disentanglement learning},
  author={Ding, Zheng and Xu, Yifan and Xu, Weijian and Parmar, Gaurav and Yang, Yang and Welling, Max and Tu, Zhuowen},
  booktitle={Proceedings of the IEEE/CVF Conference on Computer Vision and Pattern Recognition},
  pages={7920--7929},
  year={2020},
  publisher={IEEE},
  address={Virtual}
}

@inproceedings{esmaeili2019structured,
  title={Structured disentangled representations},
  author={Esmaeili, Babak and Wu, Hao and Jain, Sarthak and Bozkurt, Alican and Siddharth, Narayanaswamy and Paige, Brooks and Brooks, Dana H and Dy, Jennifer and Meent, Jan-Willem},
  booktitle={The 22nd International Conference on Artificial Intelligence and Statistics},
  pages={2525--2534},
  year={2019},
  organization={PMLR},
  publisher={PMLR},
  address={Naha, Okinawa, Japan}
}

@inproceedings{kulkarni2015deep,
  author = {Kulkarni, Tejas D and Whitney, William F. and Kohli, Pushmeet and Tenenbaum, Josh},
  booktitle = {Advances in Neural Information Processing Systems},
  editor = {C. Cortes and N. Lawrence and D. Lee and M. Sugiyama and R. Garnett},
  pages = {},
  publisher = {Curran Associates, Inc.},
  title = {Deep Convolutional Inverse Graphics Network},
  volume = {28},
  year = {2015},
  address={Montreal, Canada}
}

@inproceedings{aumentado2019geometric,
  title={Geometric disentanglement for generative latent shape models},
  author={Aumentado-Armstrong, Tristan and Tsogkas, Stavros and Jepson, Allan and Dickinson, Sven},
  booktitle={Proceedings of the IEEE/CVF International Conference on Computer Vision},
  pages={8181--8190},
  year={2019},
  publisher = {IEEE},
  address = {Seoul, Korea (South)}
}

@inproceedings{wang2021self,
  title={Self-Supervised Learning Disentangled Group Representation as Feature},
  author={Wang, Tan and Yue, Zhongqi and Huang, Jianqiang and Sun, Qianru and Zhang, Hanwang},
  booktitle={Advances in Neural Information Processing Systems},
  volume={34},
  year={2021},
  publisher = {Curran Associates, Inc.},
  address={Virtual},
  pages={}
}

@inproceedings{rhodes2021local,
  title={Local Disentanglement in Variational Auto-Encoders Using Jacobian $ L\_1 $ Regularization},
  author={Rhodes, Travers and Lee, Daniel},
  booktitle={Advances in Neural Information Processing Systems},
  volume={34},
  year={2021},
  publisher = {Curran Associates, Inc.},
  address={Virtual},
  pages={}
}

@inproceedings{cosmo2020limp,
  title="LIMP: Learning Latent Shape Representations with Metric Preservation Priors",
  author={Cosmo, Luca and Norelli, Antonio and Halimi, Oshri and Kimmel, Ron and Rodola, Emanuele},
  booktitle={European Conference on Computer Vision -- ECCV 2020},
  pages={19--35},
  year={2020},
  organization={Springer},
  publisher="Springer International Publishing",
  address="Online",
  isbn="978-3-030-58580-8"
}

@inproceedings{abrevaya2019decoupled,
	address = {Seoul, Korea (South)},
	title = {A {Decoupled} {3D} {Facial} {Shape} {Model} by {Adversarial} {Training}},
	isbn = {978-1-72814-803-8},
	language = {en},
	urldate = {2022-09-17},
	booktitle = {2019 {IEEE}/{CVF} {International} {Conference} on {Computer} {Vision} ({ICCV})},
	publisher = {IEEE},
	author = {Abrevaya, Victoria Fernandez and Boukhayma, Adnane and Wuhrer, Stefanie and Boyer, Edmond},
	month = oct,
	year = {2019},
	pages = {9418--9427},
}

@misc{aumentado2021disentangling,
  author = {Aumentado-Armstrong, Tristan and Tsogkas, Stavros and Dickinson, Sven and Jepson, Allan},
  title = {Disentangling Geometric Deformation Spaces in Generative Latent Shape Models},
  publisher = {arXiv},
  year = {2021},
  copyright = {Creative Commons Attribution Non Commercial Share Alike 4.0 International}
}

@inproceedings{zhang2020learning,
  title={Learning Distribution Independent Latent Representation for 3D Face Disentanglement},
  author={Zhang, Zihui and Yu, Cuican and Li, Huibin and Sun, Jian and Liu, Feng},
  booktitle={2020 International Conference on 3D Vision (3DV)},
  pages={848--857},
  year={2020},
  publisher={IEEE},
  address="Virtual"
}

@inproceedings{lombardi2021latenthuman,
  title={LatentHuman: Shape-and-Pose Disentangled Latent Representation for Human Bodies},
  author={Lombardi, Sandro and Yang, Bangbang and Fan, Tianxing and Bao, Hujun and Zhang, Guofeng and Pollefeys, Marc and Cui, Zhaopeng},
  booktitle={2021 International Conference on 3D Vision (3DV)},
  pages={278--288},
  year={2021},
  publisher={IEEE},
  address={Virtual}
}

@inproceedings{zheng2022imface,
  title={ImFace: A Nonlinear 3D Morphable Face Model with Implicit Neural Representations},
  author={Zheng, Mingwu and Yang, Hongyu and Huang, Di and Chen, Liming},
  booktitle={Proceedings of the IEEE/CVF Conference on Computer Vision and Pattern Recognition},
  pages={20343--20352},
  year={2022},
  publisher={IEEE},
  address={New Orleans, Louisiana, USA}
}

@misc{yang2020dsm,
  title={DSM-Net: Disentangled Structured Mesh Net for Controllable Generation of Fine Geometry},
  author={Yang, Jie and Mo, Kaichun and Lai, Yu-Kun and Guibas, Leonidas J and Gao, Lin},
  journal={arXiv preprint arXiv:2008.05440},
  year={2020},
  publisher = {arXiv},
}

@article{nash2017shape,
  author = {Nash, C. and Williams, C. K. I.},
  title = {The Shape Variational Autoencoder: A Deep Generative Model of Part-Segmented 3D Objects},
  year = {2017},
  issue_date = {August 2017},
  publisher = {The Eurographs Association & John Wiley & Sons, Ltd.},
  address = {Chichester, GBR},
  volume = {36},
  number = {5},
  issn = {0167-7055},
  journal = {Computer Graphics Forum},
  month = {8},
  pages = {1–12},
  numpages = {12},
}

@inproceedings{roberts2021lsd,
  title={LSD-StructureNet: Modeling Levels of Structural Detail in 3D Part Hierarchies},
  author={Roberts, Dominic and Danielyan, Ara and Chu, Hang and Golparvar-Fard, Mani and Forsyth, David},
  booktitle={Proceedings of the IEEE/CVF International Conference on Computer Vision},
  pages={5836--5845},
  year={2021},
  publisher={IEEE},
  address={Virtual}
}

@article{li2021editvae,
  title={EditVAE: Unsupervised Parts-Aware Controllable 3D Point Cloud Shape Generation}, 
  author={Li, Shidi and Liu, Miaomiao and Walder, Christian}, 
  volume={36}, 
  number={2}, 
  journal={Proceedings of the AAAI Conference on Artificial Intelligence}, 
  year={2022}, 
  month={6}, 
  pages={1386-1394}
}

@article{loper2015smpl,
  title={SMPL: A skinned multi-person linear model},
  author={Loper, Matthew and Mahmood, Naureen and Romero, Javier and Pons-Moll, Gerard and Black, Michael J},
  journal={ACM transactions on graphics (TOG)},
  volume={34},
  number={6},
  pages={1--16},
  year={2015},
  publisher={ACM New York, NY, USA}
}

@inproceedings{osman2020star,
  author = {Osman, Ahmed A A and Bolkart, Timo and Black, Michael J.},
  title = {{STAR}: A Sparse Trained Articulated Human Body Regressor},
  booktitle = {European Conference on Computer Vision (ECCV)},
  pages = {598--613},
  year = {2020},
  publisher="Springer International Publishing",
  address={Virtual}
}

@article{tena2011interactive,
  author = {Tena, J. Rafael and De la Torre, Fernando and Matthews, Iain},
  title = {Interactive Region-Based Linear 3D Face Models},
  year = {2011},
  issue_date = {July 2011},
  publisher = {Association for Computing Machinery},
  address = {New York, NY, USA},
  volume = {30},
  number = {4},
  issn = {0730-0301},
  journal = {ACM Transactions on Graphics},
  month = {7},
  articleno = {76},
  numpages = {10},
}

@inproceedings{blanz1999morphable,
  author = {Blanz, Volker and Vetter, Thomas},
  title = {A Morphable Model for the Synthesis of 3D Faces},
  year = {1999},
  isbn = {0201485605},
  publisher = {ACM Press/Addison-Wesley Publishing Co.},
  address = {Los Angeles, California, USA},
  booktitle = {Proceedings of the 26th Annual Conference on Computer Graphics and Interactive Techniques},
  pages = {187–194},
  numpages = {8},
  series = {SIGGRAPH '99}
}

@inproceedings{ploumpis2019combining,
  title={Combining 3D Morphable Models: A Large scale Face-and-Head Model},
  author={Ploumpis, Stylianos and Wang, Haoyang and Pears, Nick and Smith, William AP and Zafeiriou, Stefanos},
  booktitle={Proceedings of the IEEE Conference on Computer Vision and Pattern Recognition},
  pages={10934--10943},
  year={2019},
  publisher={IEEE},
  address={Long Beach, California, USA}
}

@article{ploumpis2020towards,
  title={Towards a Complete 3D Morphable Model of the Human Head}, 
  author={Ploumpis, Stylianos and Ververas, Evangelos and Sullivan, Eimear O' and Moschoglou, Stylianos and Wang, Haoyang and Pears, Nick and Smith, William A. P. and Gecer, Baris and Zafeiriou, Stefanos},
  journal={IEEE Transactions on Pattern Analysis and Machine Intelligence}, 
  year={2021},
  volume={43},
  number={11},
  pages={4142-4160},
}

@article{li2017learning,
  title={Learning a model of facial shape and expression from 4D scans.},
  author={Li, Tianye and Bolkart, Timo and Black, Michael J and Li, Hao and Romero, Javier},
  journal={ACM Trans. Graph.},
  volume={36},
  number={6},
  pages={194--1},
  year={2017}
}

@inproceedings{gong2019spiralnet++,
  title={Spiralnet++: A fast and highly efficient mesh convolution operator},
  author={Gong, Shunwang and Chen, Lei and Bronstein, Michael and Zafeiriou, Stefanos},
  booktitle={Proceedings of the IEEE/CVF International Conference on Computer Vision Workshops},
  year={2019},
  publisher={IEEE},
  address={Seoul, Korea (South)}
}

@inproceedings{ranjan2018generating,
  title={Generating 3D faces using convolutional mesh autoencoders},
  author={Ranjan, Anurag and Bolkart, Timo and Sanyal, Soubhik and Black, Michael J},
  booktitle={Proceedings of the European Conference on Computer Vision (ECCV)},
  pages={704--720},
  year={2018},
  publisher="Springer International Publishing",
  address={Munich, Germany}
}

@misc{cheng2019meshgan,
  title = {MeshGAN: Non-linear 3D Morphable Models of Faces},
  author = {Cheng, Shiyang and Bronstein, Michael and Zhou, Yuxiang and Kotsia, Irene and Pantic, Maja and Zafeiriou, Stefanos},
  publisher = {arXiv},
  year = {2019},
  copyright = {Creative Commons Attribution Non Commercial Share Alike 4.0 International}
}

@inproceedings{gecer2020synthesizing,
  title={Synthesizing coupled 3d face modalities by trunk-branch generative adversarial networks},
  author={Gecer, Baris and Lattas, Alexandros and Ploumpis, Stylianos and Deng, Jiankang and Papaioannou, Athanasios and Moschoglou, Stylianos and Zafeiriou, Stefanos},
  booktitle={European Conference on Computer Vision},
  pages={415--433},
  year={2020},
  organization={Springer},
  publisher={IEEE},
  address={Virtual}
}

@inproceedings{li2020learning,
  title={Learning Formation of Physically-Based Face Attributes},
  author={Li, Ruilong and Bladin, Karl and Zhao, Yajie and Chinara, Chinmay and Ingraham, Owen and Xiang, Pengda and Ren, Xinglei and Prasad, Pratusha and Kishore, Bipin and Xing, Jun and others},
  booktitle={Proceedings of the IEEE/CVF Conference on Computer Vision and Pattern Recognition},
  pages={3410--3419},
  year={2020},
  publisher={CVPR},
  address={Virtual}
}

@inproceedings{booth20163d,
  title={A 3d morphable model learnt from 10,000 faces},
  author={Booth, James and Roussos, Anastasios and Zafeiriou, Stefanos and Ponniah, Allan and Dunaway, David},
  booktitle={Proceedings of the IEEE Conference on Computer Vision and Pattern Recognition},
  pages={5543--5552},
  year={2016},
  publisher = {IEEE},
  address={Las Vegas, Nevada}
}

@article{dai2020statistical,
  title={Statistical modeling of craniofacial shape and texture},
  author={Dai, Hang and Pears, Nick and Smith, William and Duncan, Christian},
  journal={International Journal of Computer Vision},
  volume={128},
  number={2},
  pages={547--571},
  year={2020},
  publisher={Springer}
}

@inproceedings{yuan2020mesh,
  title={Mesh variational autoencoders with edge contraction pooling},
  author={Yuan, Yu-Jie and Lai, Yu-Kun and Yang, Jie and Duan, Qi and Fu, Hongbo and Gao, Lin},
  booktitle={Proceedings of the IEEE/CVF Conference on Computer Vision and Pattern Recognition Workshops},
  pages={274--275},
  year={2020},
  publisher={IEEE},
  address={Virtual}
}

@article{zhou2020fully,
  title={Fully convolutional mesh autoencoder using efficient spatially varying kernels},
  author={Zhou, Yi and Wu, Chenglei and Li, Zimo and Cao, Chen and Ye, Yuting and Saragih, Jason and Li, Hao and Sheikh, Yaser},
  journal={Advances in Neural Information Processing Systems},
  volume={33},
  pages={9251--9262},
  year={2020}
}

@misc{kingma2013auto,
  author = {Kingma, Diederik P and Welling, Max},
  title = {Auto-Encoding Variational Bayes},
  publisher = {arXiv},
  year = {2013},
  copyright = {arXiv.org perpetual, non-exclusive license}
}

@inproceedings{arjovsky2017wasserstein,
  title={Wasserstein generative adversarial networks},
  author={Arjovsky, Martin and Chintala, Soumith and Bottou, L{\'e}on},
  booktitle = 	 {Proceedings of the 34th International Conference on Machine Learning},
  pages = 	 {214--223},
  year = 	 {2017},
  editor = 	 {Precup, Doina and Teh, Yee Whye},
  volume = 	 {70},
  series = 	 {Proceedings of Machine Learning Research},
  month = 	 {8},
  publisher =    {PMLR},
  address = {Sydney, Australia}
}

@inproceedings{mao2017least,
  title={Least squares generative adversarial networks},
  author={Mao, Xudong and Li, Qing and Xie, Haoran and Lau, Raymond YK and Wang, Zhen and Paul Smolley, Stephen},
  booktitle={Proceedings of the IEEE international conference on computer vision},
  pages={2794--2802},
  year={2017},
  address={Venice, Italy},
  publisher={IEEE}
}

@article{pons2015dyna,
  title={Dyna: A model of dynamic human shape in motion},
  author={Pons-Moll, Gerard and Romero, Javier and Mahmood, Naureen and Black, Michael J},
  journal={ACM Transactions on Graphics (TOG)},
  volume={34},
  number={4},
  pages={1--14},
  year={2015},
  publisher={ACM New York, NY, USA}
}

@inproceedings{varol2017learning,
  title={Learning from synthetic humans},
  author={Varol, Gul and Romero, Javier and Martin, Xavier and Mahmood, Naureen and Black, Michael J and Laptev, Ivan and Schmid, Cordelia},
  booktitle={Proceedings of the IEEE Conference on Computer Vision and Pattern Recognition},
  pages={109--117},
  year={2017},
  publisher={IEEE},
  address={Honolulu, Hawaii, USA}
}

@inproceedings{zuffi20173d,
  title={3D menagerie: Modeling the 3D shape and pose of animals},
  author={Zuffi, Silvia and Kanazawa, Angjoo and Jacobs, David W and Black, Michael J},
  booktitle={Proceedings of the IEEE conference on computer vision and pattern recognition},
  pages={6365--6373},
  year={2017},
  publisher={IEEE},
  address = {Honolulu, Hawaii, USA}
}

@article{lyon2014normal,
  title={Why are normal distributions normal?},
  author={Lyon, Aidan},
  journal={The British Journal for the Philosophy of Science},
  volume={65},
  number={3},
  pages={621--649},
  year={2014},
  publisher={Oxford University Press}
}

@inproceedings{chandran2020semantic,
  title={Semantic deep face models},
  author={Chandran, Prashanth and Bradley, Derek and Gross, Markus and Beeler, Thabo},
  booktitle={2020 International Conference on 3D Vision (3DV)},
  pages={345--354},
  year={2020},
  publisher={IEEE},
  address = {Fukuoka, Japan}
}

@inproceedings{yang2019pointflow,
  title={Pointflow: 3d point cloud generation with continuous normalizing flows},
  author={Yang, Guandao and Huang, Xun and Hao, Zekun and Liu, Ming-Yu and Belongie, Serge and Hariharan, Bharath},
  booktitle={Proceedings of the IEEE/CVF International Conference on Computer Vision},
  pages={4541--4550},
  year={2019},
  publisher={IEEE},
  address={Seoul, Korea (South)}
}

@inproceedings{achlioptas2018learning,
  title={Learning representations and generative models for 3d point clouds},
  author={Achlioptas, Panos and Diamanti, Olga and Mitliagkas, Ioannis and Guibas, Leonidas},
  booktitle = 	 {Proceedings of the 35th International Conference on Machine Learning},
  pages = 	 {40--49},
  year = 	 {2018},
  editor = 	 {Dy, Jennifer and Krause, Andreas},
  volume = 	 {80},
  series = 	 {Proceedings of Machine Learning Research},
  month = 	 {7},
  publisher =    {PMLR},
  address={Stockholm, Sweden},
}

@misc{kumar2017variational,
  title={Variational Inference of Disentangled Latent Concepts from Unlabeled Observations},
  author={Kumar, Abhishek and Sattigeri, Prasanna and Balakrishnan, Avinash},
  booktitle={International Conference on Learning Representations},
  year={2018},
  publisher={OpenReview.net},
  address={Vancouver, Canada}
}

@article{claes2018genome,
  title={Genome-wide mapping of global-to-local genetic effects on human facial shape},
  author={Claes, Peter and Roosenboom, Jasmien and White, Julie D and Swigut, Tomek and Sero, Dzemila and Li, Jiarui and Lee, Myoung Keun and Zaidi, Arslan and Mattern, Brooke C and Liebowitz, Corey and others},
  journal={Nature genetics},
  volume={50},
  number={3},
  pages={414--423},
  year={2018},
  publisher={Nature Publishing Group}
}

@article{o2022convolutional,
  title={Convolutional mesh autoencoders for the 3-dimensional identification of FGFR-related craniosynostosis},
  author={O’Sullivan, Eimear and van de Lande, Lara S and Papaioannou, Athanasios and Breakey, Richard WF and Jeelani, N Owase and Ponniah, Allan and Duncan, Christian and Schievano, Silvia and Khonsari, Roman H and Zafeiriou, Stefanos and others},
  journal={Scientific reports},
  volume={12},
  number={1},
  pages={1--8},
  year={2022},
  publisher={Nature Publishing Group}
}

@article{lewis2014practice,
  title={Practice and theory of blendshape facial models.},
  author={Lewis, John P and Anjyo, Ken and Rhee, Taehyun and Zhang, Mengjie and Pighin, Frederic H and Deng, Zhigang},
  journal={Eurographics (State of the Art Reports)},
  volume={1},
  number={8},
  pages={2},
  year={2014}
}

@inproceedings{litany2018deformable,
  title={Deformable shape completion with graph convolutional autoencoders},
  author={Litany, Or and Bronstein, Alex and Bronstein, Michael and Makadia, Ameesh},
  booktitle={Proceedings of the IEEE conference on computer vision and pattern recognition},
  pages={1886--1895},
  year={2018},
  publisher={IEEE},
  address={Salt Lake City, Utah, USA}
}

@inproceedings{yang2018foldingnet,
  title={Foldingnet: Point cloud auto-encoder via deep grid deformation},
  author={Yang, Yaoqing and Feng, Chen and Shen, Yiru and Tian, Dong},
  booktitle={Proceedings of the IEEE conference on computer vision and pattern recognition},
  pages={206--215},
  year={2018},
  publisher={IEEE},
  address={Salt Lake City, Utah, USA}
}

@inproceedings{dai2019pointae,
  title={Pointae: Point auto-encoder for 3d statistical shape and texture modelling},
  author={Dai, Hang and Shao, Ling},
  booktitle={Proceedings of the IEEE/CVF International Conference on Computer Vision},
  pages={5410--5419},
  year={2019},
  publisher={IEEE},
  address={Seoul, Korea (South)}
}

@article{tan2022mesh,
  title={Variational Autoencoders for Localized Mesh Deformation Component Analysis}, 
  author={Tan, Qingyang and Zhang, Ling-Xiao and Yang, Jie and Lai, Yu-Kun and Gao, Lin},
  journal={IEEE Transactions on Pattern Analysis and Machine Intelligence}, 
  year={2022},
  volume={44},
  number={10},
  pages={6297-6310},
}

@inproceedings{bouritsas2019neural,
  title={Neural 3d morphable models: Spiral convolutional networks for 3d shape representation learning and generation},
  author={Bouritsas, Giorgos and Bokhnyak, Sergiy and Ploumpis, Stylianos and Bronstein, Michael and Zafeiriou, Stefanos},
  booktitle={Proceedings of the IEEE/CVF International Conference on Computer Vision},
  pages={7213--7222},
  year={2019},
  publisher = {IEEE},
  address = {Seoul, Korea (South)}
}

@misc{olivier2021facetunegan,
  title={FaceTuneGAN: Face Autoencoder for Convolutional Expression Transfer Using Neural Generative Adversarial Networks},
  author={Olivier, Nicolas and Baert, Kelian and Danieau, Fabien and Multon, Franck and Avril, Quentin},
  journal={arXiv preprint arXiv:2112.00532},
  year={2021},
  publisher = {arXiv},
  copyright = {arXiv.org perpetual, non-exclusive license}
}

@inproceedings{zhou2020unsupervised,
  title={Unsupervised shape and pose disentanglement for 3d meshes},
  author={Zhou, Keyang and Bhatnagar, Bharat Lal and Pons-Moll, Gerard},
  booktitle={European Conference on Computer Vision},
  pages={341--357},
  year={2020},
  organization={Springer},
  publisher="Springer International Publishing",
  address={Virtual}
}

@misc{
  tatro2021unsupervised,
  title={Unsupervised Geometric Disentanglement via {CFAN}-{VAE}},
  author={Norman Joseph Tatro and Stefan C Schonsheck and Rongjie Lai},
  booktitle={ICLR 2021 Workshop on Geometrical and Topological Representation Learning},
  year={2021},
  publisher={OpenReview.net},
  address={Virtual}
}

@inproceedings{jiang2019disentangled,
  title={Disentangled representation learning for 3d face shape},
  author={Jiang, Zi-Hang and Wu, Qianyi and Chen, Keyu and Zhang, Juyong},
  booktitle={Proceedings of the IEEE/CVF Conference on Computer Vision and Pattern Recognition},
  pages={11957--11966},
  year={2019},
  publisher={IEEE},
  address={Long Beach, California, USA}
}

@inproceedings{huang2021arapreg,
  title={ARAPReg: An As-Rigid-As Possible Regularization Loss for Learning Deformable Shape Generators},
  author={Huang, Qixing and Huang, Xiangru and Sun, Bo and Zhang, Zaiwei and Jiang, Junfeng and Bajaj, Chandrajit},
  booktitle={Proceedings of the IEEE/CVF International Conference on Computer Vision},
  pages={5815--5825},
  year={2021},
  publisher={IEEE},
  address={Virtual}
}

@inproceedings{shoshan2021gan,
  title={Gan-control: Explicitly controllable gans},
  author={Shoshan, Alon and Bhonker, Nadav and Kviatkovsky, Igor and Medioni, Gerard},
  booktitle={Proceedings of the IEEE/CVF International Conference on Computer Vision},
  pages={14083--14093},
  year={2021},
  publisher={IEEE},
  address={Virtual}
}

@misc{huang2021multimodal,
  title={Multimodal Conditional Image Synthesis with Product-of-Experts GANs},
  author={Huang, Xun and Mallya, Arun and Wang, Ting-Chun and Liu, Ming-Yu},
  journal={arXiv preprint arXiv:2112.05130},
  year={2021},
  publisher = {arXiv},
  copyright = {arXiv.org perpetual, non-exclusive license}
}

@inproceedings{karras2021alias,
  title={Alias-free generative adversarial networks},
  author={Karras, Tero and Aittala, Miika and Laine, Samuli and H{\"a}rk{\"o}nen, Erik and Hellsten, Janne and Lehtinen, Jaakko and Aila, Timo},
  journal={Advances in Neural Information Processing Systems},
  volume={34},
  year={2021},
  booktitle = {Advances in Neural Information Processing Systems},
  pages = {852--863},
  publisher = {Curran Associates, Inc.},
}

@article{harkonen2020ganspace,
  title={Ganspace: Discovering interpretable gan controls},
  author={H{\"a}rk{\"o}nen, Erik and Hertzmann, Aaron and Lehtinen, Jaakko and Paris, Sylvain},
  journal={Advances in Neural Information Processing Systems},
  volume={33},
  pages={9841--9850},
  year={2020}
}

@inproceedings{alharbi2020disentangled,
  title={Disentangled image generation through structured noise injection},
  author={Alharbi, Yazeed and Wonka, Peter},
  booktitle={Proceedings of the IEEE/CVF Conference on Computer Vision and Pattern Recognition},
  pages={5134--5142},
  year={2020},
  address = {Virtual},
  publisher = {IEEE}
}

@article{shen2020interfacegan,
  title={InterFaceGAN: Interpreting the Disentangled Face Representation Learned by GANs}, 
  author={Shen, Yujun and Yang, Ceyuan and Tang, Xiaoou and Zhou, Bolei},
  publisher={IEEE},
  journal={IEEE Transactions on Pattern Analysis and Machine Intelligence}, 
  year={2022},
  volume={44},
  number={4},
  pages={2004-2018},
}

@inproceedings{voynov2020unsupervised,
  title={Unsupervised discovery of interpretable directions in the gan latent space},
  author={Voynov, Andrey and Babenko, Artem},
  booktitle={International conference on machine learning},
  pages={9786--9796},
  year={2020},
  organization={PMLR},
  publisher={PMLR},
  address={Virtual}
}

@article{he2019attgan,
  title={Attgan: Facial attribute editing by only changing what you want},
  author={He, Zhenliang and Zuo, Wangmeng and Kan, Meina and Shan, Shiguang and Chen, Xilin},
  journal={IEEE transactions on image processing},
  volume={28},
  number={11},
  pages={5464--5478},
  year={2019},
  publisher={IEEE}
}

@inproceedings{lee2020maskgan,
  title={Maskgan: Towards diverse and interactive facial image manipulation},
  author={Lee, Cheng-Han and Liu, Ziwei and Wu, Lingyun and Luo, Ping},
  booktitle={Proceedings of the IEEE/CVF Conference on Computer Vision and Pattern Recognition},
  pages={5549--5558},
  year={2020},
  publisher={IEEE},
  address={Virtual}
}

@inproceedings{ling2021editgan,
  title={EditGAN: High-Precision Semantic Image Editing},
  author={Ling, Huan and Kreis, Karsten and Li, Daiqing and Kim, Seung Wook and Torralba, Antonio and Fidler, Sanja},
  booktitle = {Advances in Neural Information Processing Systems},
  pages = {16331--16345},
  publisher = {Curran Associates, Inc.},
  volume = {34},
  year = {2021},
  address={Virtual}
}

@inproceedings{chen2021intrinsic,
  title={Intrinsic-extrinsic preserved gans for unsupervised 3D pose transfer},
  author={Chen, Haoyu and Tang, Hao and Shi, Henglin and Peng, Wei and Sebe, Nicu and Zhao, Guoying},
  booktitle={Proceedings of the IEEE/CVF International Conference on Computer Vision},
  pages={8630--8639},
  year={2021},
  publisher={IEEE},
  address={Virtual}
}

@article{li2021sp,
  title={SP-GAN: Sphere-guided 3D shape generation and manipulation},
  author={Li, Ruihui and Li, Xianzhi and Hui, Ka-Hei and Fu, Chi-Wing},
  journal={ACM Transactions on Graphics (TOG)},
  volume={40},
  number={4},
  pages={1--12},
  year={2021},
  publisher={ACM New York, NY, USA}
}

@book{chavel1984eigenvalues,
  title={Eigenvalues in Riemannian geometry},
  author={Chavel, Isaac},
  year={1984},
  publisher={Academic press},
  address={Orlando, Florida}
}

@inproceedings{defferrard2016convolutional,
  author = {Defferrard, Micha\"{e}l and Bresson, Xavier and Vandergheynst, Pierre},
  title = {Convolutional Neural Networks on Graphs with Fast Localized Spectral Filtering},
  year = {2016},
  publisher = {Curran Associates Inc.},
  address = {Red Hook, NY, USA},
  booktitle = {Proceedings of the 30th International Conference on Neural Information Processing Systems},
  pages = {3844–3852},
  numpages = {9},
  location = {Barcelona, Spain},
  series = {NIPS'16}
}

@misc{bruna2013spectral,
  title = {Spectral Networks and Locally Connected Networks on Graphs},
  author={Bruna, Joan and Zaremba, Wojciech and Szlam, Arthur and LeCun, Yann},
  publisher = {arXiv},
  year = {2013},
  copyright = {arXiv.org perpetual, non-exclusive license}
}

@inproceedings{zhu2020learning,
  title={Learning disentangled representations with latent variation predictability},
  author={Zhu, Xinqi and Xu, Chang and Tao, Dacheng},
  booktitle={European Conference on Computer Vision},
  pages={684--700},
  year={2020},
  publisher="Springer International Publishing",
  address={Virtual}
}

@article{gao2014compact,
  title={A compact shape descriptor for triangular surface meshes},
  author={Gao, Zhanheng and Yu, Zeyun and Pang, Xiaoli},
  journal={Computer-Aided Design},
  volume={53},
  pages={62--69},
  year={2014},
  publisher={Elsevier}
}

@article{reuter2006laplace,
  title={Laplace--Beltrami spectra as ‘Shape-DNA’of surfaces and solids},
  author={Reuter, Martin and Wolter, Franz-Erich and Peinecke, Niklas},
  journal={Computer-Aided Design},
  volume={38},
  number={4},
  pages={342--366},
  year={2006},
  publisher={Elsevier}
}

@InProceedings{radford21a,
  title = 	 {Learning Transferable Visual Models From Natural Language Supervision},
  author =       {Radford, Alec and Kim, Jong Wook and Hallacy, Chris and Ramesh, Aditya and Goh, Gabriel and Agarwal, Sandhini and Sastry, Girish and Askell, Amanda and Mishkin, Pamela and Clark, Jack and Krueger, Gretchen and Sutskever, Ilya},
  booktitle = 	 {Proceedings of the 38th International Conference on Machine Learning},
  pages = 	 {8748--8763},
  year = 	 {2021},
  editor = 	 {Meila, Marina and Zhang, Tong},
  volume = 	 {139},
  series = 	 {Proceedings of Machine Learning Research},
  publisher =    {PMLR},
  address={Virtual}
}

@article{egger20203d,
  title={3d morphable face models—past, present, and future},
  author={Egger, Bernhard and Smith, William AP and Tewari, Ayush and Wuhrer, Stefanie and Zollhoefer, Michael and Beeler, Thabo and Bernard, Florian and Bolkart, Timo and Kortylewski, Adam and Romdhani, Sami and others},
  journal={ACM Transactions on Graphics (TOG)},
  volume={39},
  number={5},
  pages={1--38},
  year={2020},
  publisher={ACM New York, NY, USA}
}

@article{moschoglou20203dfacegan,
  title={3dfacegan: Adversarial nets for 3d face representation, generation, and translation},
  author={Moschoglou, Stylianos and Ploumpis, Stylianos and Nicolaou, Mihalis A and Papaioannou, Athanasios and Zafeiriou, Stefanos},
  journal={International Journal of Computer Vision},
  volume={128},
  pages={2534--2551},
  year={2020},
  publisher={Springer}
}

@article{shuman2013emerging,
  title={The emerging field of signal processing on graphs: Extending high-dimensional data analysis to networks and other irregular domains},
  author={Shuman, David I and Narang, Sunil K and Frossard, Pascal and Ortega, Antonio and Vandergheynst, Pierre},
  journal={IEEE signal processing magazine},
  volume={30},
  number={3},
  pages={83--98},
  year={2013},
  publisher={IEEE}
}

@inproceedings{wang2019backpropagation,
  title={Backpropagation-friendly eigendecomposition},
  author={Wang, Wei and Dang, Zheng and Hu, Yinlin and Fua, Pascal and Salzmann, Mathieu},
  booktitle={Advances in Neural Information Processing Systems},
  volume={32},
  year={2019},
  address={Virtual}
}

@article{marin2021spectral,
  title={Spectral shape recovery and analysis via data-driven connections},
  author={Marin, Riccardo and Rampini, Arianna and Castellani, Umberto and Rodol{\`a}, Emanuele and Ovsjanikov, Maks and Melzi, Simone},
  journal={International journal of computer vision},
  volume={129},
  pages={2745--2760},
  year={2021},
  publisher={Springer}
}

@inproceedings{otberdout2022sparse,
  title={Sparse to dense dynamic 3d facial expression generation},
  author={Otberdout, Naima and Ferrari, Claudio and Daoudi, Mohamed and Berretti, Stefano and Del Bimbo, Alberto},
  booktitle={Proceedings of the IEEE/CVF Conference on Computer Vision and Pattern Recognition},
  pages={20385--20394},
  year={2022},
  publisher={IEEE},
  address={New Orleans, Louisiana, USA}
}

\newpage

\end{document}


\maketitle

\section{PCA-Based Baseline}
\begin{figure}[b]
        \centering
        \includegraphics[width=\linewidth]{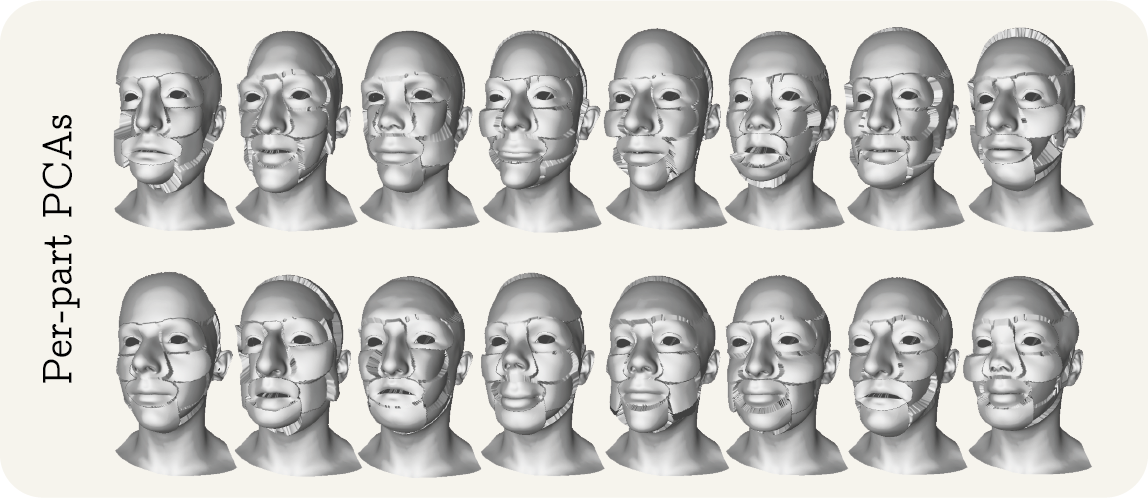}
        \caption{Random samples generated by a bundle of PCA models trained each on a different shape attribute.}
        \label{a-fig:pca}
    \end{figure}

    LED models and SD-VAE are characterised by a single architecture that, thanks to a disentangled latent representation, can control the generation of local shape attributes while still considering the whole output shape. To demonstrate the need for these models we compare our method also against a bundle of attribute-specific PCA models. As it can be observed in Fig.~\ref{a-fig:pca}, the main issue of naive per-part methods, such as the bundle of PCA models, is that shape attributes are independently generated. Even though this makes the different attributes fully disentangled between each others, significant surface discontinuities appear during the generation procedure. On the contrary, LED models are capable of ensuring the continuity of the output surfaces while providing control over the generation of each attribute.
    
\section{Mesh Operators}
    \label{a-sec:mesh-operators}
    Traditional neural network operators are not well suited for non-Euclidean data such as meshes. In recent years, many operators capable of operating on meshes were proposed. We decide to build all models with the intuitive spiral++ convolutions and with quadric-sampling-based pooling operators as in [FKSC22; GCBZ19]. However, other mesh operators could be used. Spiral++ convolutions, which are specifically designed to efficiently operate on datasets of meshes sharing the same topology [GCBZ19], are built aggregating vertices along spiral sequences and processing them with a multilayer perceptron. Spirals are defined for each vertex of the mesh by selecting an arbitrary neighbour as well as the other vertices along a clockwise spiral. All spirals are precomputed and have a fixed length. The receptive field of these convolutions can be increased by dilating the spirals, thus skipping a predefined amount of vertices after each selected vertex. Since spirals are precomputed, only the multilayer preceptron's weights are learned during training. Also the pooling operators are precomputed. In fact, a quadric sampling procedure that iteratively contracts the vertex pair with the smallest quadric error is applied to the mean shape of the training data $\mathbf{M}$. During this procedure both a pooling and an un-pooling sparse matrix are defined. The former has values of $1$ in correspondence of vertices that need to be preserved and $0$ elsewhere. The latter still has values of $1$ for vertices that remain unchanged while un-pooling, but it also stores the barycentric weights corresponding to the barycentric coordinates of contracted vertices in order to restore them. These precomputed sparse matrices are matrix multiplied with the vertex features computed by the different network's layers to achieve pooling and un-pooling. 

\section{Architectures}
    \label{a-sec:architectures}
    
    \paragraph{VAEs operating on meshes from \textsc{Uhm}.} 
        The architecture of encoder and generator are defined as: 
        \begin{equation*}
            \begin{split}
                E \! = \! e(\text{Conv}(32)) \xrightarrow{\downarrow 4} e(\text{Conv}(32)) \xrightarrow{\downarrow 4} e(\text{Conv}(32)) \\ \xrightarrow{\downarrow 4} e(\text{Conv}(64)) \xrightarrow{\downarrow 4} 2\times \text{Lin}(60)   
            \end{split}
        \end{equation*}
        \begin{equation*}
            \begin{split}
                G \! = \! \text{Lin}(64 n) \xrightarrow{\uparrow 4} e(\text{Conv}(64)) \xrightarrow{\uparrow 4} e(\text{Conv}(32)) \\ \xrightarrow{\uparrow 4} e(\text{Conv}(32)) \xrightarrow{\uparrow 4} e(\text{Conv}(32)) \rightarrow \text{Conv}(3),  
            \end{split}
        \end{equation*}
        where Lin($\cdot$) and Conv($\cdot$) respectively represent linear layers and spiral convolutions (Sec.~\ref{a-sec:mesh-operators}) with their number of output features. $e(\cdot)$ is the ELU (exponential linear unit) non-linear activation function. Right arrows represent pooling operations (Sec.~\ref{a-sec:mesh-operators}). Their superscript indicates the sampling factor as well as whether it is an up-sampling~($\uparrow$) or a down-sampling~($\downarrow$) operator. $n$ is the number of down-sampled vertices after all the sampling operations. In this case, being $N$ the number of input and output vertices, $n = N/4^4$. Note that $E$ terminates with two linear layers that are responsible to predict $\bm{\mu}$ and $\bm{\sigma}$, which are used in conjunction with the reparametrization trick to sample a latent vector $\mathbf{z}$. This vector is the input to the generator $G$.
        
        The architecture described above is used by VAE, LED-VAE, SD-VAE, and DIP-VAE-I.
        
    \paragraph{VAEs operating on meshes from \textsc{Lyhm} and Co\textsc{ma}.}
        Since meshes from \textsc{Lyhm} and Co\textsc{ma} have fewer vertices than \textsc{Uhm}, the pooling layers of the VAE models have smaller sampling factor. Therefore, we have:
        \begin{equation*}
            \begin{split}
                E \! = \! e(\text{Conv}(32)) \xrightarrow{\downarrow 4} e(\text{Conv}(32)) \xrightarrow{\downarrow 2} e(\text{Conv}(32)) \\ \xrightarrow{\downarrow 2} e(\text{Conv}(64)) \xrightarrow{\downarrow 2} 2\times \text{Lin}(60)  
            \end{split}
        \end{equation*}
        \begin{equation*}
            \begin{split}
                G \! = \! \text{Lin}(64 n) \xrightarrow{\uparrow 2} e(\text{Conv}(64)) \xrightarrow{\uparrow 2} e(\text{Conv}(32)) \\ \xrightarrow{\uparrow 2} e(\text{Conv}(32)) \xrightarrow{\uparrow 4} e(\text{Conv}(32)) \rightarrow \text{Conv}(3).  
            \end{split}
        \end{equation*}
        
    \paragraph{VAEs operating on meshes from \textsc{Star}.}
        Also \textsc{Star} has fewer vertices than \textsc{Uhm}. In this case we use the same architecture used by [FKSC22]. on this dataset:
        \begin{equation*}
            E \! = \! e(\text{Conv}(32)) \xrightarrow{\downarrow 4} e(\text{Conv}(32)) \xrightarrow{\downarrow 4} e(\text{Conv}(64)) \xrightarrow{\downarrow 4} 2\times \text{Lin}(60)   
        \end{equation*}
        \begin{equation*}
            \begin{split}
                G \! = \! \text{Lin}(64 n) \xrightarrow{\uparrow 4} e(\text{Conv}(64)) \xrightarrow{\uparrow 4} e(\text{Conv}(32)) \\ \xrightarrow{\uparrow 4} e(\text{Conv}(32)) \rightarrow \text{Conv}(3).  
            \end{split}
        \end{equation*}
        
    \paragraph{LSGANs operating on meshes from \textsc{Uhm}.}
        As mentioned in Sec.~3, the architecture of $G$ remains the same of the VAEs operating on meshes from \textsc{Uhm} and the architecture of the discriminator $D$ is similar to the one of $E$ with some minor difference: 
        \begin{equation*}
            \begin{split}
                G \! = \! \text{Lin}(64 n) \xrightarrow{\uparrow 4} e(\text{Conv}(64)) \xrightarrow{\uparrow 4} e(\text{Conv}(32)) \\ \xrightarrow{\uparrow 4} e(\text{Conv}(32)) \xrightarrow{\uparrow 4} e(\text{Conv}(32)) \rightarrow \text{Conv}(3)  
            \end{split}
        \end{equation*}
        \begin{equation*}
            \begin{split}
                D \! = \! e(\text{Conv}(32)) \xrightarrow{\downarrow 4} e(\text{Conv}(32)) \xrightarrow{\downarrow 4} e(\text{Conv}(32)) \\ \xrightarrow{\downarrow 4} e(\text{Conv}(64)) \xrightarrow{\downarrow 4} \text{Lin}(60). 
            \end{split}
        \end{equation*}
        Note that the main difference between $E$ and $D$ is the number of linear layers at the end of the architecture. This architecture is used for both LSGAN and LED-LSGAN.
        
    \paragraph{WGANs operating on meshes from \textsc{Uhm}.}
        The architecture of the generator in WGAN is the same as LSGAN's. The architecture of the critic $C$ is the same as $D$ with fewer neurons in the last linear layer, which outputs a single value. In fact, we have: 
        \begin{equation*}
            \begin{split}
                G \! = \! \text{Lin}(64 n) \xrightarrow{\uparrow 4} e(\text{Conv}(64)) \xrightarrow{\uparrow 4} e(\text{Conv}(32)) \\ \xrightarrow{\uparrow 4} e(\text{Conv}(32)) \xrightarrow{\uparrow 4} e(\text{Conv}(32)) \rightarrow \text{Conv}(3)      
            \end{split}
        \end{equation*}
        \begin{equation*}
            \begin{split}
                C \! = \! e(\text{Conv}(32)) \xrightarrow{\downarrow 4} e(\text{Conv}(32)) \xrightarrow{\downarrow 4} e(\text{Conv}(32)) \\ \xrightarrow{\downarrow 4} e(\text{Conv}(64)) \xrightarrow{\downarrow 4} \text{Lin}(1). 
            \end{split}
        \end{equation*}
        This architecture is used for both WGAN and LED-WGAN.
        
        \begin{figure*}[ht]
            \centering
            \includegraphics[width=\textwidth]{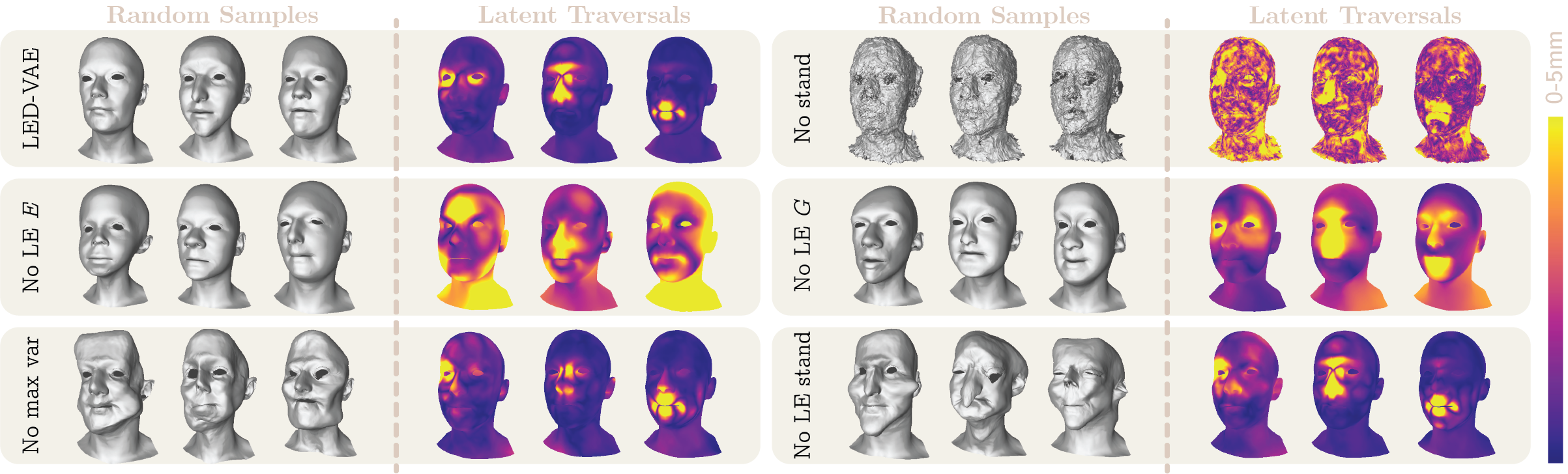}
            \caption{Ablation study. The LED-VAE is ablated by removing the data standardisation and the computation of the local eigenprojection loss on either the encoder or generator. Albations are performed also selecting the first eigenvectors instead of those associated with the maximum variance and without standardising the local eigenprojections in the loss computation.}
            \label{a-fig:ablation}
        \end{figure*}
        
\section{Implementation Details}
    \label{a-sec:implementation}
    All networks are implemented in PyTorch using the mesh operators described in Sec.~\ref{a-sec:mesh-operators} with the implementation made available by [FKSC22]~\footnote{The SD-VAE, DIP-VAE, and FactorVAE implementations as well as the evaluation code are publicly available at \href{https://github.com/simofoti/3DVAE-SwapDisentangled}{github.com/simofoti/3DVAE-SwapDisentangled}}. 
    We segment heads from \textsc{Uhm}, \textsc{Lyhm}, and Co\textsc{ma} in $F=12$ head attributes and split the latent representation $\mathbf{z}$ in $F$ subsets of size $\kappa=5$. Bodies from \textsc{Star} have $F=11$ and $\kappa=3$.
    For the sake of comparison, all models trained on the meshes obtained from \textsc{Uhm}, Co\textsc{ma}, and \textsc{Star} are trained for $40$ epochs. Since \textsc{Lyhm} has significantly fewer meshes, models trained on this dataset are trained for $400$ epochs, which correspond to approximately the same number of iterations as other models. In addition, the batch size is always set to $16$ while spirals length and dilation are set to $9$ and $1$ respectively. 
    Data are always standardised by subtracting the per-vertex mean of the training set ($\mathbf{M}$) and dividing by the per-vertex standard deviation of the training set ($\mathbf{\Sigma}$). 
    
    \paragraph{VAEs.}
        All VAE models are trained with the \textsc{Adam} optimizer using a fixed learning rate of $1e^{-4}$ and a KL divergence weight set to $\beta = 1e^{-4}$. The vanilla VAE, DIP-VAE-I, and SD-VAE have a smoothing loss weight of $\alpha=1$. As reported in [FKSC22], the latent consistency weight of the SD-VAE model is set to $1$ and the contrastive margins to $0.5$. In DIP-VAE-I we set $\lambda_d=100$ and $\lambda_{od}=10$.
        
    \paragraph{LED-VAEs.}    
        During the eigendecomposition of the Kirchoff Laplacians $\mathbf{K}_\omega$ we compute the first $K=50$ eigenvectors. The $\alpha$ weight controlling the smoothing loss is set to $\alpha=50$, while the weights of the local eigenprojection losses are set to $\eta_1=1$ and $\eta_2=0.5$. As previously mentioned, the weight controlling the KL divergence is set to $\beta = 1e^{-4}$.
        The values reported above are used when LED-VAE is trained on the meshes from \textsc{Uhm}. When LED-VAE is trained on \textsc{Lyhm}, we have: $K=45$, $\alpha=10$, $\beta=1e^{-4}$, $\eta_1=0.5$ and $\eta_2=0.25$. On Co\textsc{ma} we have: $K=45$, $\alpha=50$, $\beta=1e^{-4}$, $\eta_1=1$ and $\eta_2=2$. On \textsc{Star} we have: $K=50$, $\alpha=10$, $\beta=1e^{-4}$, $\eta_1=0.1$ and $\eta_2=2$.

    \paragraph{LSGAN and LED-LSGAN.}
        The generator $G$ is trained using the \textsc{Adam} optimizer with a fixed learning rate of $1e^{-4}$. The discriminator $D$ using SGD with a fixed learning rate of $8e^{-4}$. The weight of the Laplacian smoothing term is set to $\alpha=10$ in LSGAN and $\alpha=50$ in LED-LSGAN. In LED-LSGAN the local eigenprojection loss weight is set to $\eta=0.5$ and $K=50$ eigenvalues are computed during the eigendecompositions of the $\mathbf{K}_\omega$.
        
    \paragraph{WGAN and LED-WGAN.}
        Both generator and critc are trained using the RMSprop optimizer. The learning rate of the optimizer operating on $G$ is set to $1e^{-4}$, while the learning rate of the optimizer operating on $C$ is $5e^{-5}$. The $C$ network weights are clipped to the range $[-c, c]$ with $c=0.01$. The weight of the Laplacian smoothing term is set to $\alpha=10$ in WGAN and $\alpha=50$ in LED-WGAN. In LED-WGAN the local eigenprojection loss weight is set to $\eta=0.25$ and $K=50$ eigenvalues are computed.

\section{LED-VAE Additional Experiments}
    As mentioned in Sec.~5, we consider LED-VAE to be the most promising generative model among the proposed LED models because it is simpler to train, requires less hyperparameter tuning, and generates higher quality meshes. Therefore, we conduct experiments to evaluate the importance of the different assumptions made in its construction (Sec.~\ref{a-subsec:ablation} and Sec.~\ref{a-subsec:segmentations}), and observe the smoothness of the latent space (Sec.~\ref{a-subsec:interpolations}). 

        \begin{figure}[b]
            \centering
            \includegraphics[width=\linewidth]{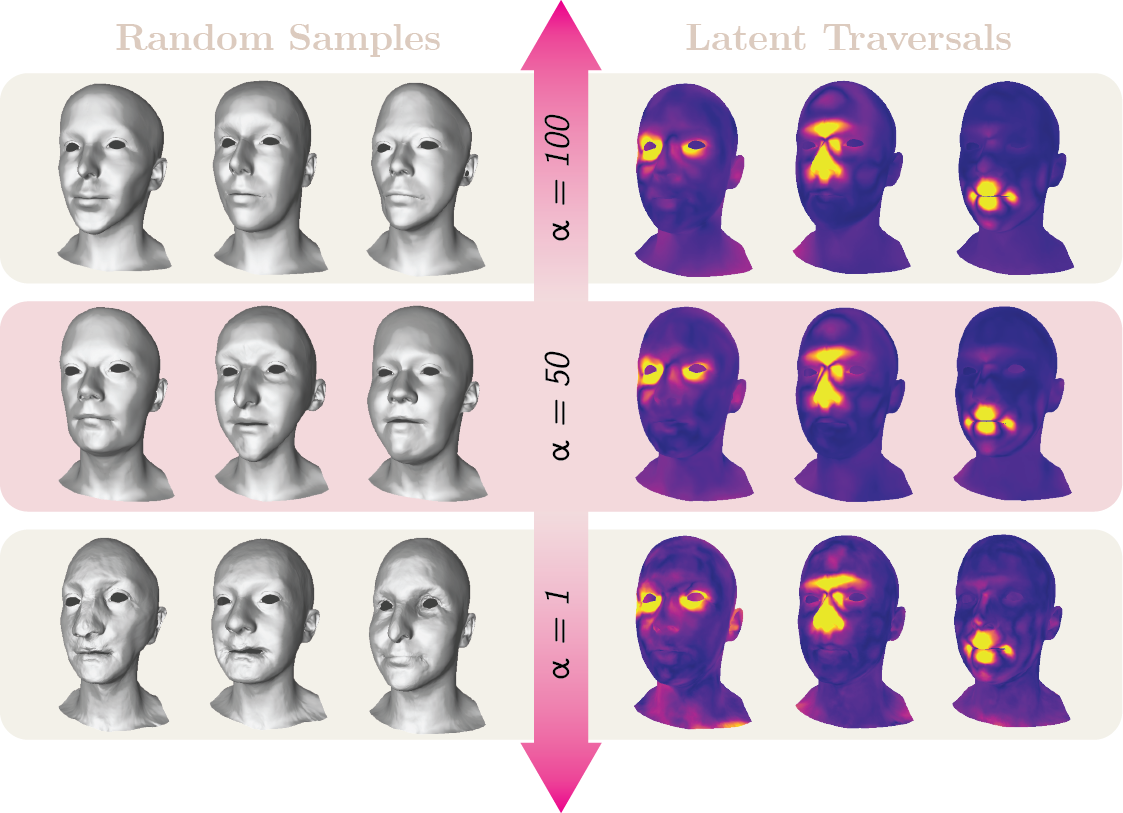}
            \caption{Effects of smoothness weight ($\alpha$) on random samples and latent traversals. The row highlighted in pink reports results obtained with the proposed implementation of LED-VAE. Latent traversals are referred always to the same latent variable.}
                \label{a-fig:alpha-ablation}
        \end{figure}

        \begin{figure*}
            \makebox[\textwidth][c]{\includegraphics[width=\textwidth]{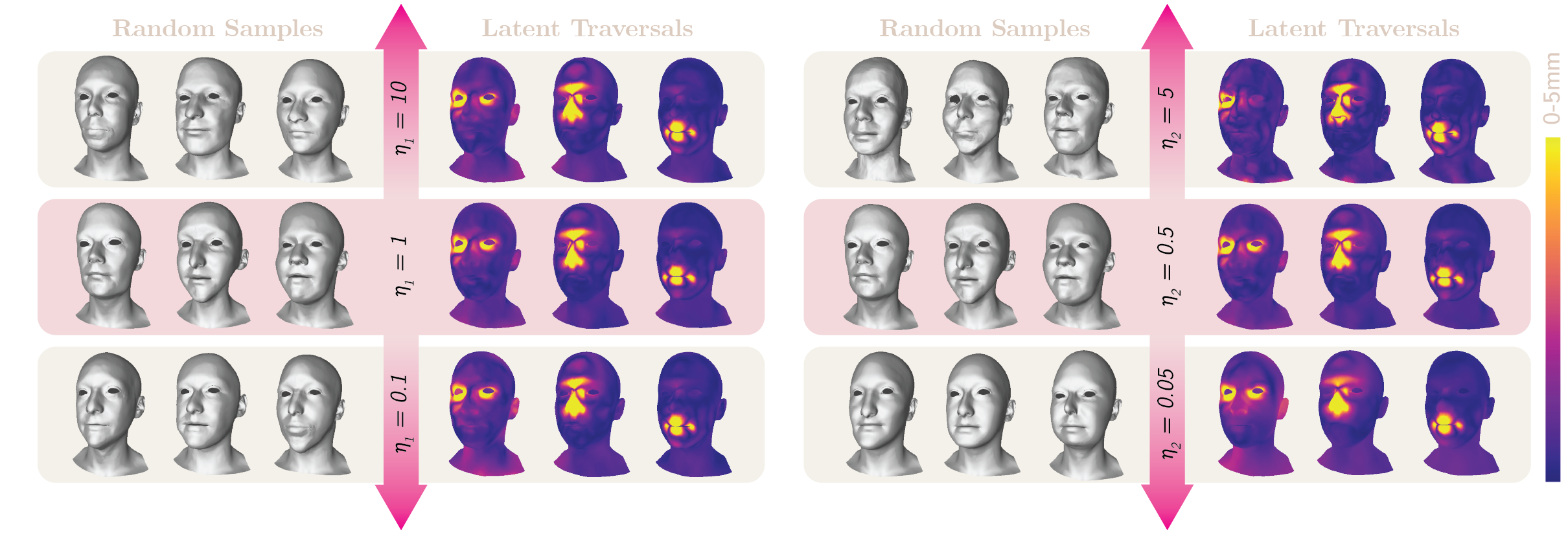}}
            \centering
            \caption{Effects of local eigenprojection weights on random samples and latent traversals. Rows highlighted in pink report results obtained with the proposed implementation of LED-VAE. The left column shows the effects of changing $\eta_1$, which controls the local eigenprojection loss affecting the encoder. The right column the effects of changing $\eta_2$, which controls the local eigenprojection loss affecting the generator. Latent traversals are referred always to the same latent variable.}
            \label{a-fig:eta-ablation}
        \end{figure*}
        
    \subsection{Ablation Study}
        \label{a-subsec:ablation}
        
        The ablation study in Fig.~\ref{a-fig:ablation} is performed by re-training the proposed LED-VAE without some of its characterising design choices. Models are re-trained with the same architecture (Sec.~\ref{a-sec:architectures}) and implementation details (Sec.~\ref{a-sec:implementation}) of LED-VAE. Only one design choice is altered per ablation experiment. Not standardising the data (No stand in Fig.~\ref{a-fig:ablation}) we observe noisy random samples. Some control over the generation of local attributes appears to be retained, but the presence of noise contributes to shape variations across the entire shape during latent traversals. When the local eigenprojection loss is not computed on the encoder ($\eta_1 = 0$), not only the encoder loses its disentanglement capabilities, but the control over the generation of local shape attributes is significantly reduced (No LE $E$ in Fig.~\ref{a-fig:ablation}). Similar results are obtained when the local eigenprojection loss is not computed on the generator ($\eta_2 = 0$), though the encoder should retain its disentanglement power (No LE $G$ in Fig.~\ref{a-fig:ablation}). If instead of selecting the $\kappa$ eigenvectors corresponding to the spectral components with the highest variance, we select the first $\kappa$ eigenvectors as Fourier modes for the local eigenprojection, the generator creates unrealistic shapes (No max var in Fig.~\ref{a-fig:ablation}). Unrealistic shapes are generated also if the local eigenprojections are not standardised and thus $\mathbf{m}_\omega^\star = 0$ and $\mathbf{s}_\omega^\star = 1$ in Eq.~$3$ (No LE stand in Fig.~\ref{a-fig:ablation}). In addition, note that the vanilla VAE is equivalent to the LED-VAE without local eigenprojection losses and its results are equivalent to those of an ablation experiment where both the local eigenprojection losses are set to zero.

        Not only we perform an ablation study removing some characterising design choice of LED-VAE, but we also experiment with the strength of their weighting coefficients. In Fig.~\ref{a-fig:alpha-ablation}, we observe the effects caused by changing the smoothing weight ($\alpha$). As expected, reducing $\alpha$ reduces also the quality of the randomly generated samples. Interestingly, also the disentanglement performance slightly deteriorates. Increasing $\alpha$ does not have major effects on the generation and disentanglement performance. In Fig.~\ref{a-fig:eta-ablation}, we report the effects of altering the local eigenprojection weighting coefficients. Even though Fig.~\ref{a-fig:ablation} (No LE E) shows the importance of enforcing the local eigenprojection loss on the encoder, we do not observe significant difference when altering its weight ($\eta_1$). Most differences can be appreciated in the first latent traversal, showing how lower weights slightly deteriorate disentanglement. On the contrary, $\eta_2$, which modulates the disentanglement on the generator, has more influence on sample quality and disentanglement. In fact, high $\eta_2$ values improve disentanglement, but reduce sample quality. 

    \subsection{Different segmentations}
        \label{a-subsec:segmentations}
        
\begin{figure*}
            \makebox[\textwidth][c]{\includegraphics[width=\textwidth]{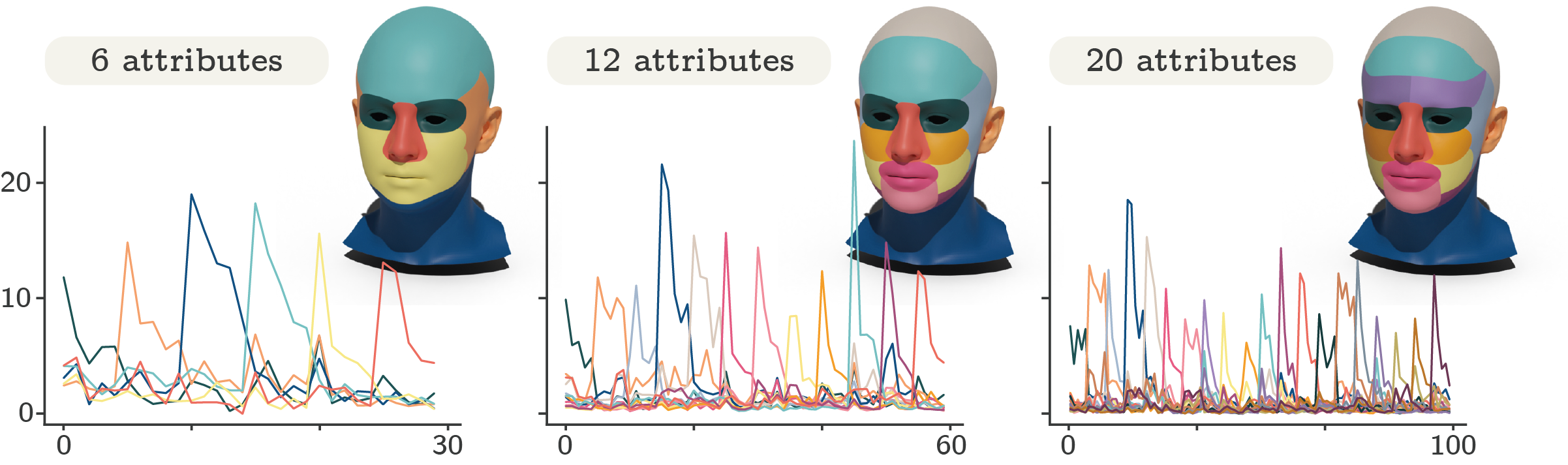}}
            \centering
            \caption{Effects of traversing each latent variable of LED-VAE trained enforcing latent disentanglement with different attribute segmentations. Note that since $5$ latent variables are used to represent each attribute, the latent size with $6$ attributes is equal to $30$, $60$ with $12$, and $100$ with $20$. When $20$ attributes are disentangled, not only we segment the supraorbital area, but we also separate the left from the right attribute. For instance, while with $12$ attributes left and right eye were grouped together, now they are separate.}
            \label{a-fig:other-segmentations}
        \end{figure*}

        \noindent The segmentations in Fig. 3 were performed with clinical supervision and were aimed at identifying key anatomical areas of the face and body. Nevertheless, different segmentations are admissible. To observe the disentanglement performance of LED-VAE with different segmentations, we re-trained LED-VAE using a coarser and a finer segmentation. As shown in Fig.~\ref{a-fig:other-segmentations}, LED-VAE successfully disentangles local identity attributes when varying the size and number of local shape attributes.

        Note that the local segments are used only by the local eigenprojection losses and are not an input to the network. For this reason, attributes can be connected, overlapping or even not connected. The segmentation used to train our models is connected. Overlapping segments could be used, but big overlaps may be counterproductive as they would increase entanglement between neighbouring regions and may produce unexpected results when eigenprojections of overlapping regions are incompatible.
        
    \subsection{Latent Interpolations and Replacements}
        \label{a-subsec:interpolations}
\begin{figure*}
            \centering
            \includegraphics[width=\textwidth]{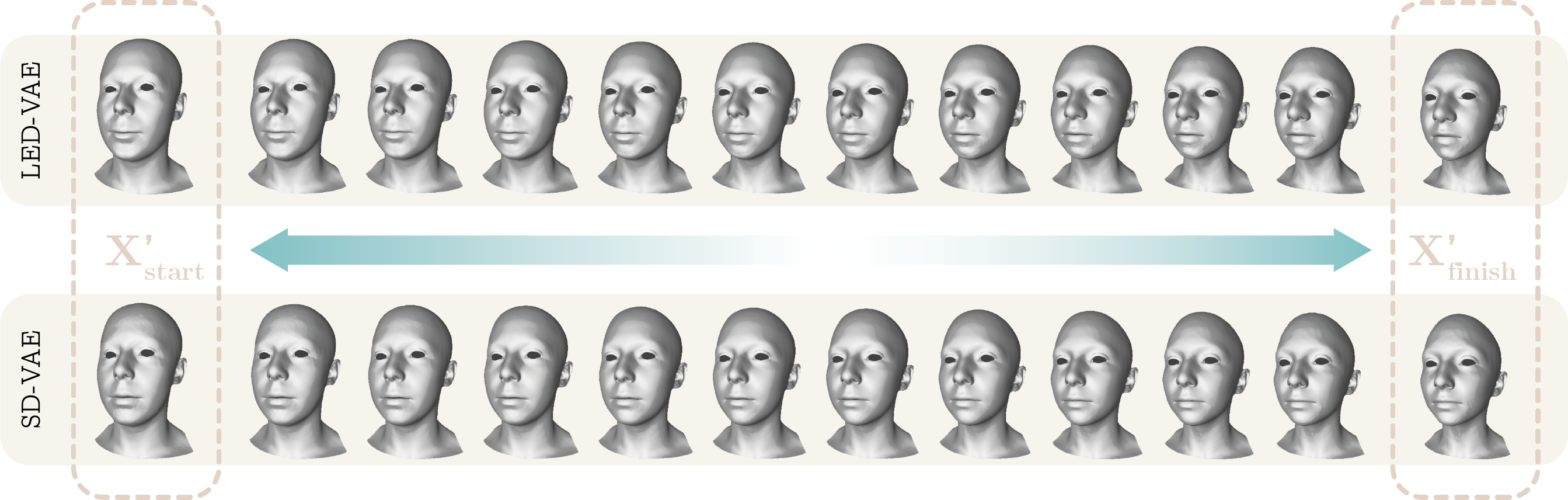}
            \caption{Latent interpolations with LED-VAE and SD-VAE. Two shapes ($\mathbf{X}_\text{start}$ and $\mathbf{X}_\text{finish}$) are randomly selected from the test set. Their latent representation is computed by feeding the two shapes in the encoder network. $10$ intermediate latent vectors are obtained by linearly interpolating all the latent variables. Shapes generated from these latent vectors smoothly transition from the reconstructed initial ($\mathbf{X'}_\text{start}$) and final shapes ($\mathbf{X'}_\text{finish}$). }
            \label{a-fig:interp-all}
        \end{figure*}

\begin{figure*}
            \centering
            \includegraphics[width=\textwidth]{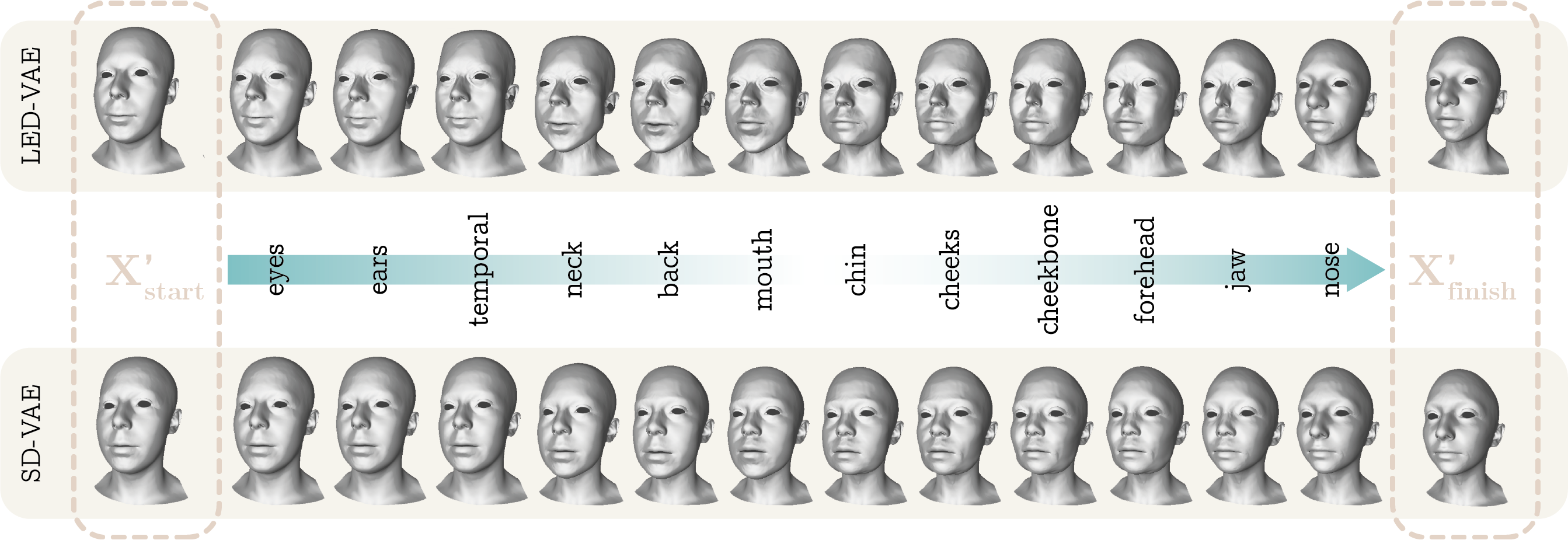}
            \caption{Per-attribute latent replacements with LED-VAE and SD-VAE. Subsets of the latent variables corresponding to different head attributes ($\mathbf{z}_\omega$) are progressively replaced. While the left-most and right-most heads are the reconstruction of the initial and target shape, the others are obtained with latent replacements. Each shape is generated starting from the one on its left. For example, the second heads from the left are generated with the latent vector of $\mathbf{X}'_\text{start}$ and replacing the subset of latent variables controlling the eyes of $\mathbf{X}'_\text{start}$ with the subset controlling the eyes of $\mathbf{X'}_\text{finish}$. Similarly, the third head has the same latent representation of the second one, but also the subset of latent variables controlling the ears is replaced. The remaining shapes are obtained repeating the same procedure.}
            \label{a-fig:interp-attributes}
        \end{figure*}

\begin{figure*}
            \centering
            \includegraphics[width=\textwidth]{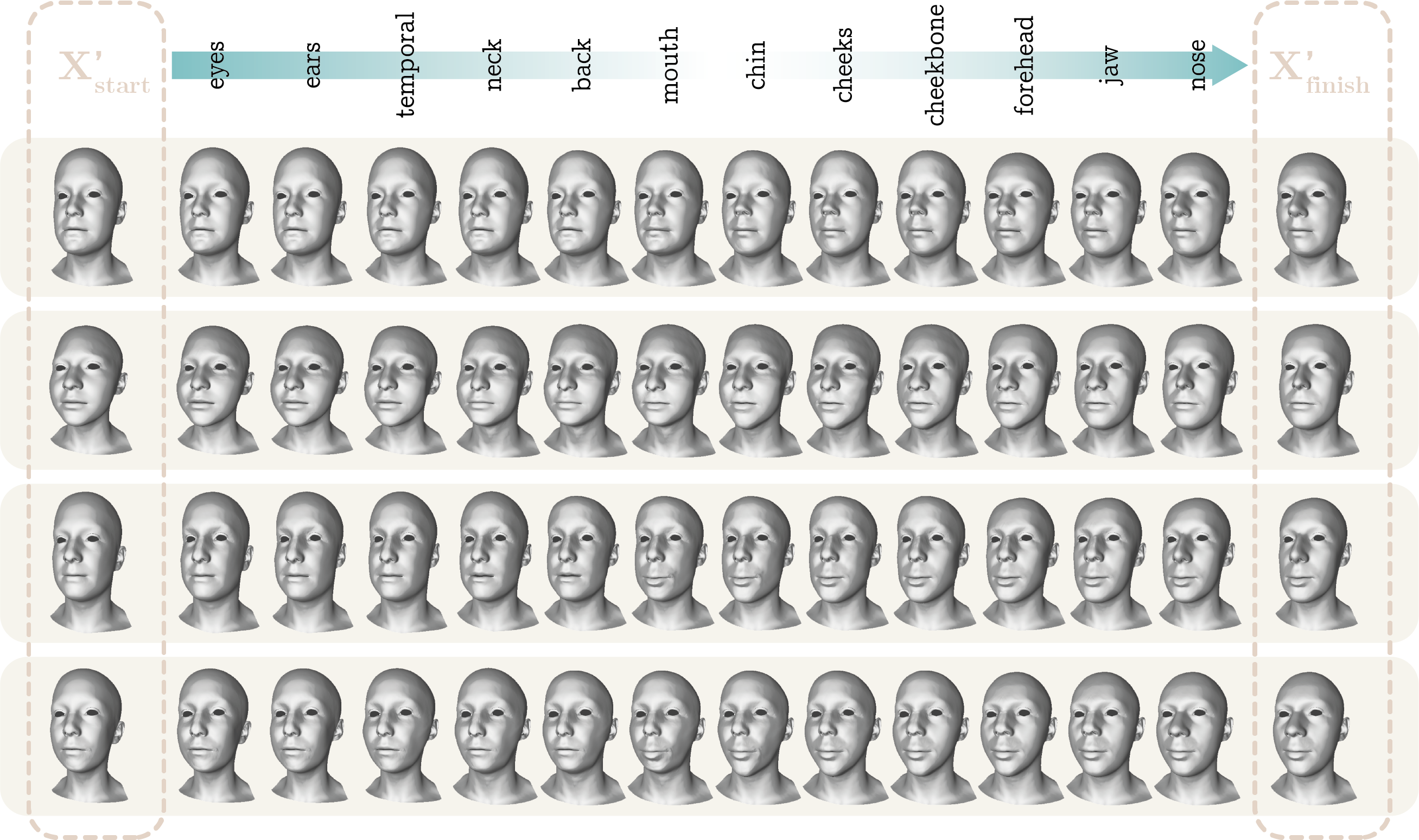}
            \caption{Additional per-attribute latent replacements with LED-VAE (see Fig.~\ref{a-fig:interp-attributes}).}
            \label{a-fig:interp-attributes-more}
        \end{figure*}

\begin{figure*}
            \centering
            \includegraphics[width=\textwidth]{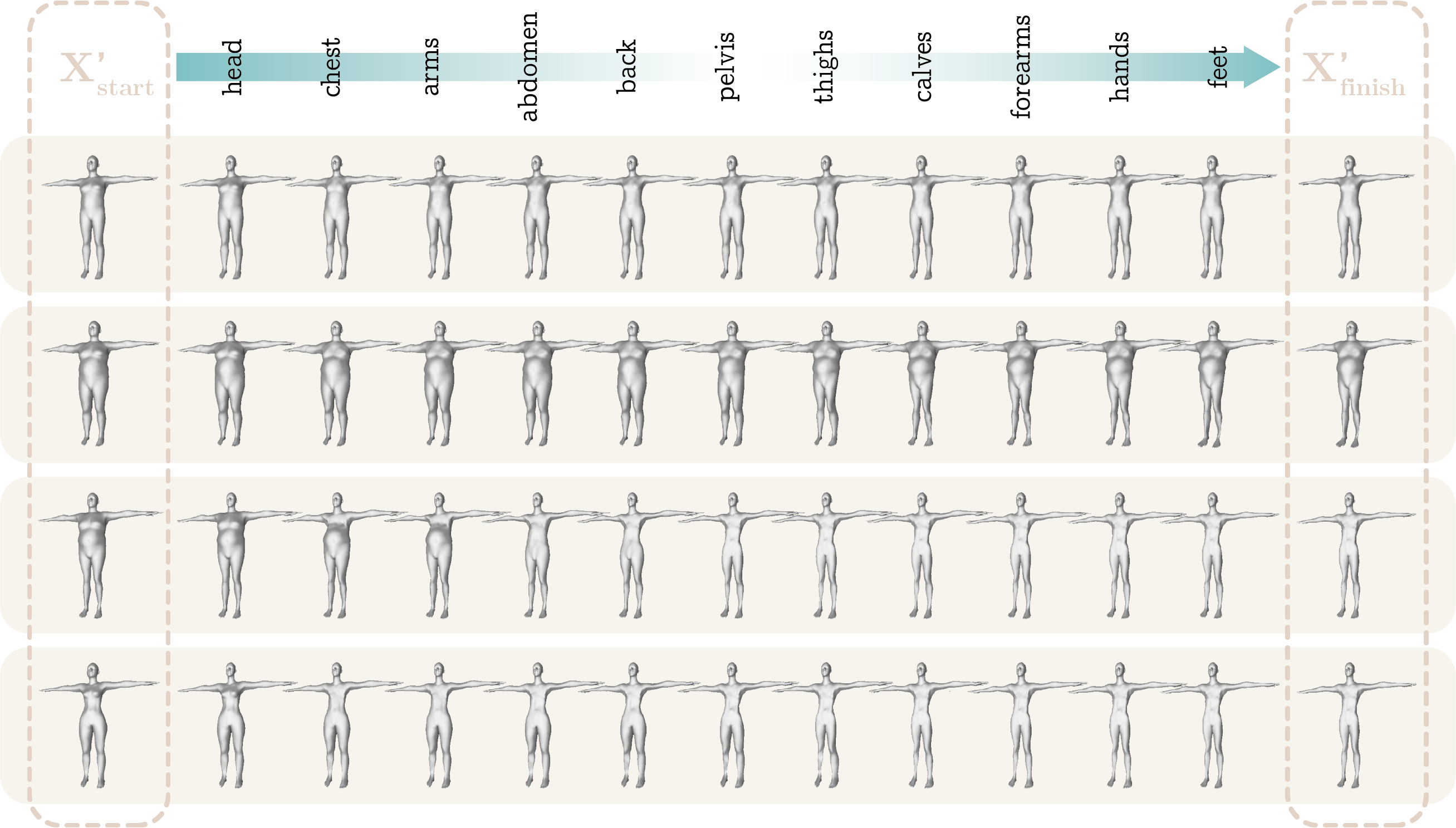}
            \caption{Additional per-attribute latent replacements with LED-VAE trained on shapes from \textsc{Star} (see Fig.~\ref{a-fig:interp-attributes}).}
            \label{a-fig:interp-attributes-body}
        \end{figure*}

        We perform two latent interpolation experiments and compare results between LED-VAE and SD-VAE, the two variational autoencoders providing control over the generation of local shape attributes. Two randomly selected test shapes $\mathbf{X}_\text{start}$ and $\mathbf{X}_\text{finish}$ are encoded to compute their respective latent representations. 
        Fig.~\ref{a-fig:interp-all} shows the reconstructed shapes $\mathbf{X}'_\text{start}$ and $\mathbf{X}'_\text{finish}$ as well as the shapes generated from latent vectors linearly interpolated between the latents of $\mathbf{X}_\text{start}$ and $\mathbf{X}_\text{finish}$. Fig.~\ref{a-fig:interp-attributes}, Fig.~\ref{a-fig:interp-attributes-more}, and Fig.~\ref{a-fig:interp-attributes-body} depict the effects of changing each $\mathbf{z}_\omega$ of $\mathbf{X}_\text{start}$ with the corresponding $\mathbf{z}_\omega$ of $\mathbf{X}_\text{finish}$. This is equivalent to progressively replacing attributes of the initial mesh with those of the final mesh. The experiment in Fig.~\ref{a-fig:interp-attributes} is better represented in the \textit{supplementary video}, where each $\mathbf{z}_\omega$ is interpolated instead of being replaced. These experiments show that the latent space of our LED-VAE is smooth. Even though some self-intersections is visible on the ears of heads generated by LED-VAE, this model appears to be better than SD-VAE at replacing attributes.
        

\section{Random Generation and Latent Disentanglement}
\begin{figure*}
        \makebox[\textwidth][c]{\includegraphics[width=\textwidth]{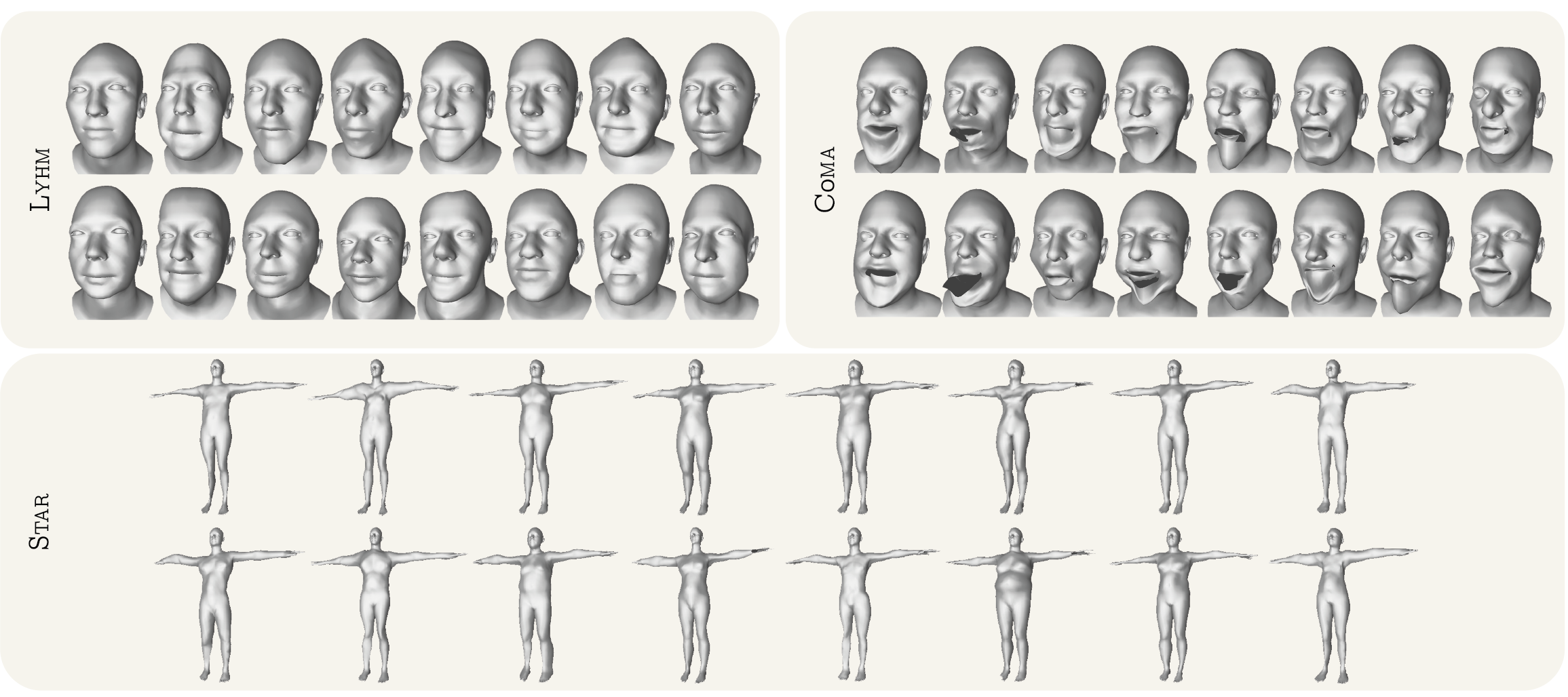}}
        \caption{Random samples generated by LED-VAE models trained on shapes from \textsc{Lyhm}, Co\textsc{ma}, and \textsc{Star}.}
        \label{a-fig:all-random-others}
    \end{figure*}

\begin{figure*}
        \makebox[\textwidth][c]{\includegraphics[width=\textwidth]{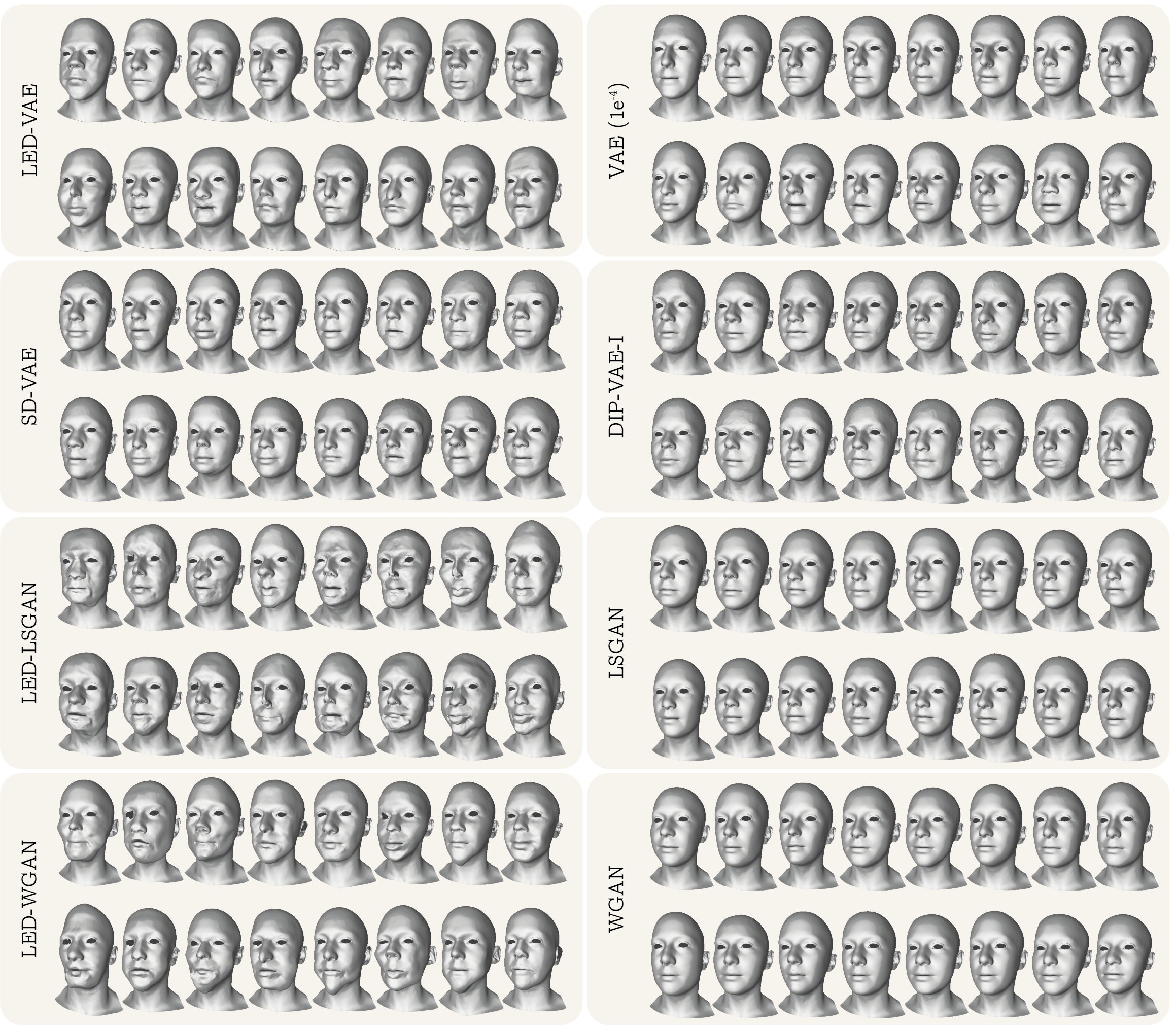}}
        \caption{Random samples generated by LED-VAE, SD-VAE, LED-LSGAN, LED-WGAN, VAE, DIP-VAE-I, LSGAN, WGAN. All models are trained on head shapes from \textsc{Uhm}.}
        \label{a-fig:all-random}
    \end{figure*}

\begin{figure*}
        \makebox[\textwidth][c]{\includegraphics[width=\textwidth]{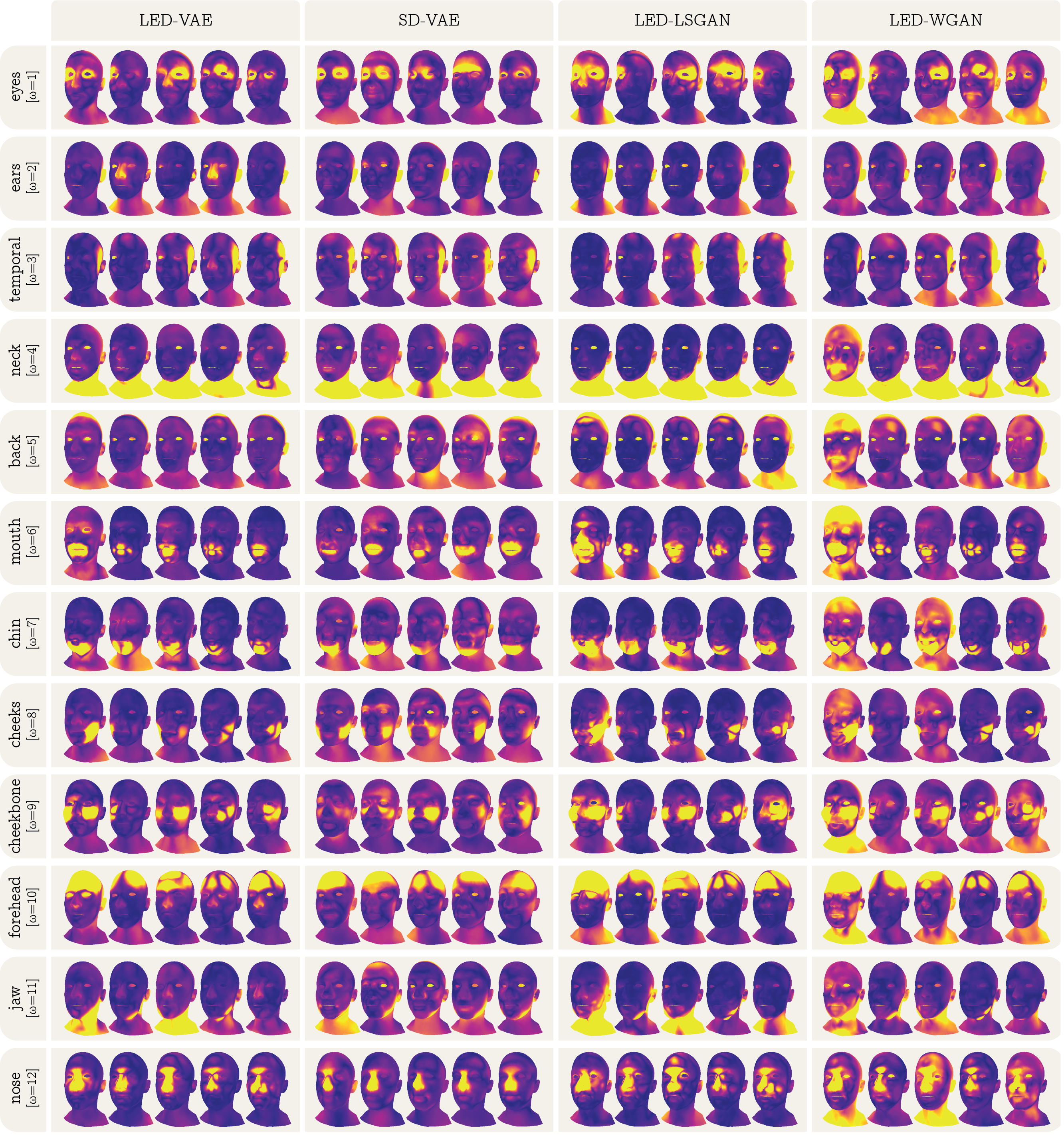}}
        \caption{Complete latent traversals grouped per-method along columns and per-attribute along rows. LED-VAE, SD-VAE, LED-LSGAN, and LED-WGAN are all trained on \textsc{Uhm}.}
        \label{a-fig:led-disentanglement}
    \end{figure*}

\begin{figure*}
        \centering
        \includegraphics[width=\textwidth]{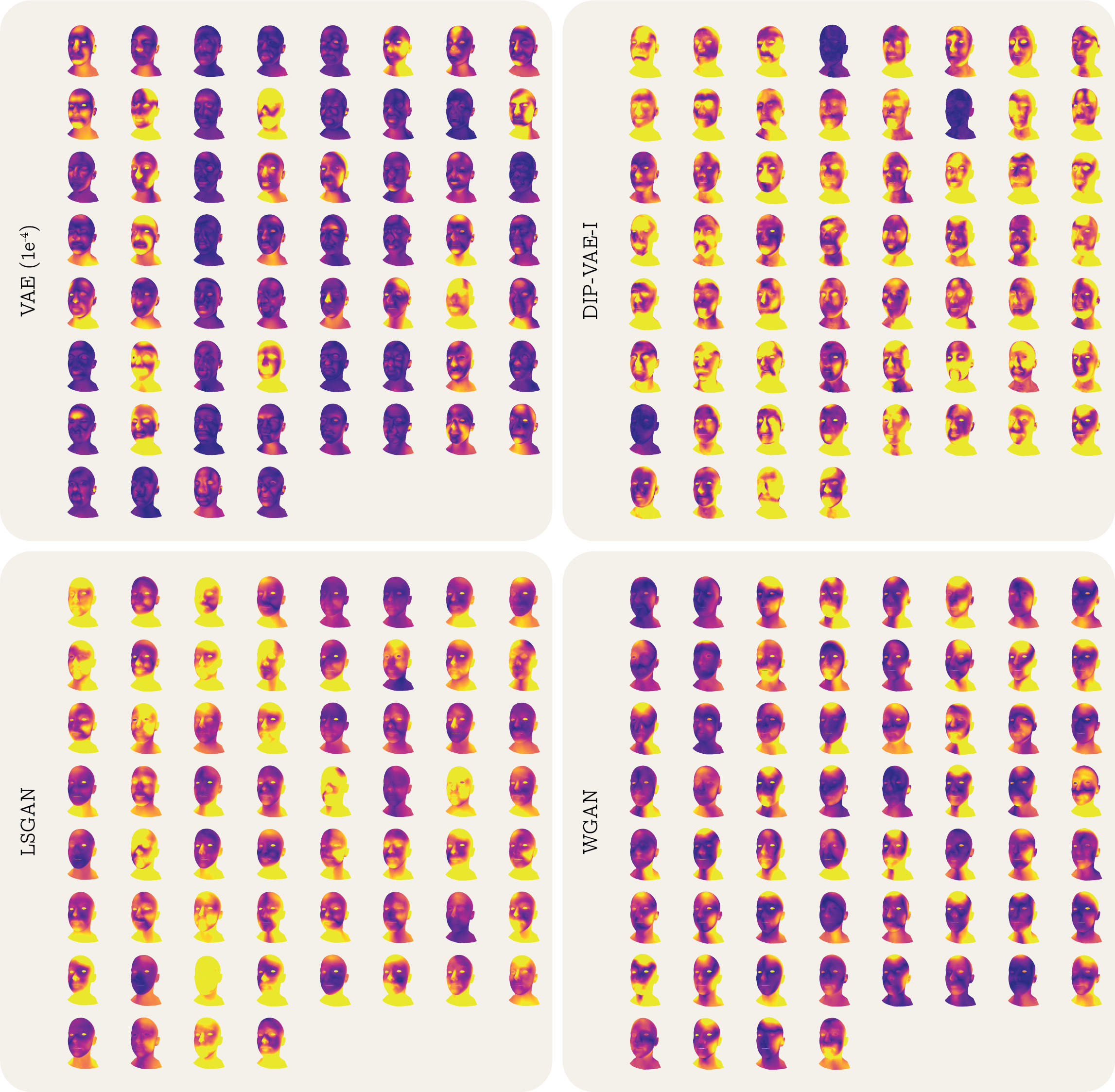}
        \caption{Complete latent traversals of VAE, DIP-VAE-I, LSGAN, and WGAN trained on \textsc{Uhm}.}
        \label{a-fig:disentanglement-comparison}
    \end{figure*}

\begin{figure*}
        \makebox[\textwidth][c]{\includegraphics[width=1\textwidth]{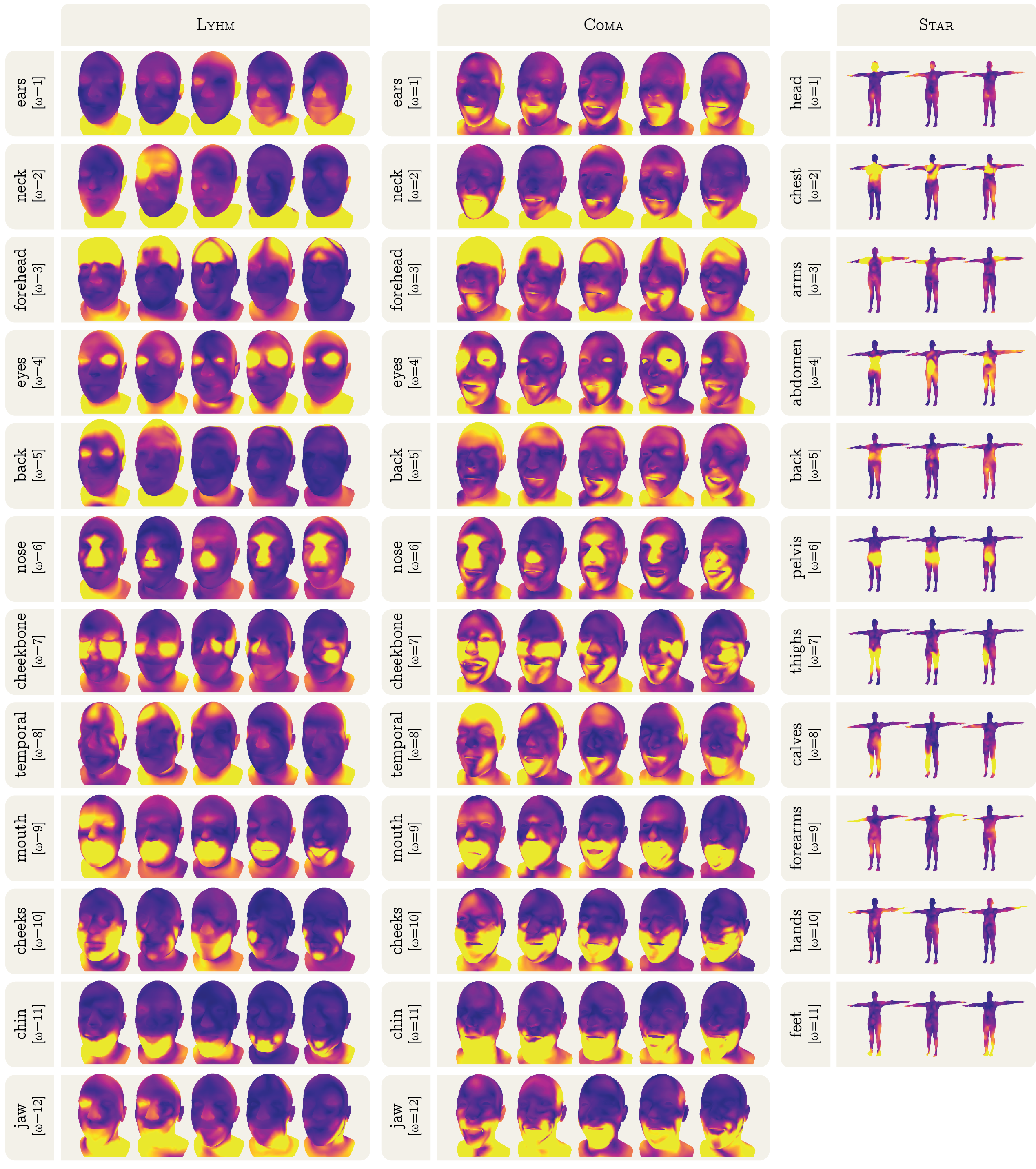}}
        \caption{Complete latent traversals of LED-VAE grouped per-dataset along columns and per-attribute along rows. The LED-VAE models are trained on shapes from \textsc{Lyhm}, Co\textsc{ma}, and \textsc{Star}.}
        \label{a-fig:led-disentanglement-others}
    \end{figure*}

    We report more randomly generated shapes and latent traversals than those already depicted in Fig.~4 and Fig.~6. In fact, in Fig.~\ref{a-fig:all-random} we show shapes obtained by all methods trained on heads from \textsc{Uhm} and in Fig.~\ref{a-fig:all-random-others} shapes generated with LED-VAE trained on \textsc{Lyhm}, and Co\textsc{ma}, and \textsc{Star}. Then we report the effects caused in the generated shapes by traversing all the latent variables. In particular, Fig.~\ref{a-fig:led-disentanglement} shows the shapes generated by traversing all $5$ latent variables in each $\mathbf{z}_\omega$ for LED-VAE, SD-VAE, LED-LSGAN, and LED-WGAN. Fig.~\ref{a-fig:disentanglement-comparison} represents the effects of latent traversals for methods that are not able to enforce disentanglement with respect to local shape attributes, such as: VAE, DIP-VAE-I, LSGAN, and WGAN. Finally, Fig.~\ref{a-fig:led-disentanglement-others} reports latent traversal results for LED-VAE trained on \textsc{Lyhm}, and Co\textsc{ma}, and \textsc{Star}.

